%% file: Full_Text_extended.tex
\documentclass[sigconf]{acmart}
\AtBeginDocument{%
  }

\setcopyright{acmlicensed}
\copyrightyear{2026}
\acmYear{2026}
\setcopyright{cc}
\setcctype{by}
\acmConference[KDD '26]{Proceedings of the 32nd ACM SIGKDD Conference on Knowledge Discovery and Data Mining V.2}{August 09--13, 2026}{Jeju Island, Republic of Korea}
\acmBooktitle{Proceedings of the 32nd ACM SIGKDD Conference on Knowledge Discovery and Data Mining V.2 (KDD '26), August 09--13, 2026, Jeju Island, Republic of Korea}
\acmDOI{10.1145/3770855.3818040}
\acmISBN{979-8-4007-2259-2/2026/08}

\usepackage{amsfonts}       
\usepackage{amsmath}        

\usepackage{amssymb}
\usepackage{bm}             
\usepackage{booktabs}       
\usepackage{multirow}       
\usepackage{graphicx}       
\usepackage{enumitem}       
\usepackage{array}          
\usepackage{tabularx}       
\usepackage{xcolor}
\usepackage{xspace}
\usepackage[utf8]{inputenc} 
\usepackage[T1]{fontenc}    
\usepackage{hyperref}       
\usepackage{url}            
\usepackage{booktabs}       
\usepackage{amsfonts}       
\usepackage{nicefrac}       
\usepackage{microtype}      
\usepackage{xcolor}         

\usepackage{multirow}
\usepackage{graphicx}
\usepackage{amsmath}
\usepackage{makecell}
\usepackage{xspace}
\usepackage{color}
\usepackage{wrapfig}
\usepackage{bbding}
\usepackage{tcolorbox}
\usepackage{enumitem}
\usepackage{amsthm}

\usepackage{array}
\usepackage{tabularx}
\usepackage{tcolorbox}
\usepackage{graphicx}
\usepackage{float}
\usepackage{placeins}
\usepackage{caption}
\usepackage{subcaption}
\usepackage{booktabs}
\usepackage{graphicx}
\usepackage{tabularx}
\newcolumntype{C}{>{\centering\arraybackslash}X}
\usepackage{bm}
\usepackage{pifont}
\usepackage{amssymb}
\usepackage{xcolor}           
\usepackage{tcolorbox}
\definecolor{myorange}{RGB}{230,97,0}
\definecolor{myblue}{RGB}{30,144,255}
\definecolor{mygreen}{RGB}{44,160,44}

\input{macro}

\begin{document}

\title{Trajectory-Aware Clinical Risk Prediction via Severity-Grounded \\ Knowledge Graphs and Retrieval-Augmented Generation}

\author{Kyunghoon Jeon}
\orcid{0009-0003-4833-0738}
\affiliation{%
  \institution{Hanyang University}
  \city{Seoul}
  \country{Republic of Korea}}
\email{kyunghoon123@hanyang.ac.kr}

\author{Youmin Ko}
\orcid{0009-0007-9369-6842}
\affiliation{%
  \institution{Hanyang University}
  \city{Seoul}
  \country{Republic of Korea}}
\email{youminkk0213@hanyang.ac.kr}

\author{Woohwan Jung}
\orcid{0000-0003-4561-2214}
\affiliation{%
  \institution{Korea University}
  \city{Seoul}
  \country{Republic of Korea}}
\email{woohwan@korea.ac.kr}

\author{Hyunjoon Kim}
\orcid{0009-0009-4019-312X}
\authornote{Corresponding author.}
\affiliation{%
  \institution{Hanyang University}
  \city{Seoul}
  \country{Republic of Korea}}
\email{hyunjoonkim@hanyang.ac.kr}

\renewcommand{\shortauthors}{Jeon et al.}

\begin{abstract}
While Electronic Health Records (EHRs) offer a wealth of clinical data, effectively augmenting a patient's records with heterogeneous external knowledge to predict the patient's clinical risk remains a significant challenge. Existing methods fail to capture disease severity, treatment responses, and nuanced clinical progression, due to data sparsity and the underutilization of unstructured clinical notes. To address these challenges, we propose \textbf{TRACER} (\textit{a trajectory-aware and clinically grounded prediction framework}) that (1) constructs a medical knowledge graph enriched with severity information from medical literature, (2) retrieves clinically relevant, severity‑weighted paths of a patient’s progression from the knowledge graph, (3) extracts clinically relevant events from unstructured clinical notes, and (4) augments patient context with similar peer cases. 
Experiments on the MIMIC‑III and MIMIC‑IV datasets demonstrate large gains over state‑of‑the‑art baselines, with up to \(28.5\%\) increase in Macro F1 score for the mortality prediction task, and \(19.7\%\) increase for the readmission prediction task.
\end{abstract}

\ccsdesc[500]{Applied computing~Health care information systems}
\ccsdesc[300]{Information systems~Information retrieval}
\ccsdesc[500]{Computing methodologies~Knowledge representation and reasoning}

\keywords{Electronic Health Records, Healthcare Outcome Prediction, Knowledge Graph, Severity Scoring, Retrieval-Augmented Generation}


\maketitle

\newcommand\kddavailabilityurl{https://doi.org/10.1145/3770855.3818040}
\ifdefempty{\kddavailabilityurl}{}{
\begingroup\small\noindent\raggedright\textbf{Resource Availability:}\\
The source code of this paper has been made publicly available at \url{https://github.com/KyunghoonJeon/TRACER.git}.
\endgroup
}

\begin{figure}[t]
    \centering
    \includegraphics[width=1\columnwidth]{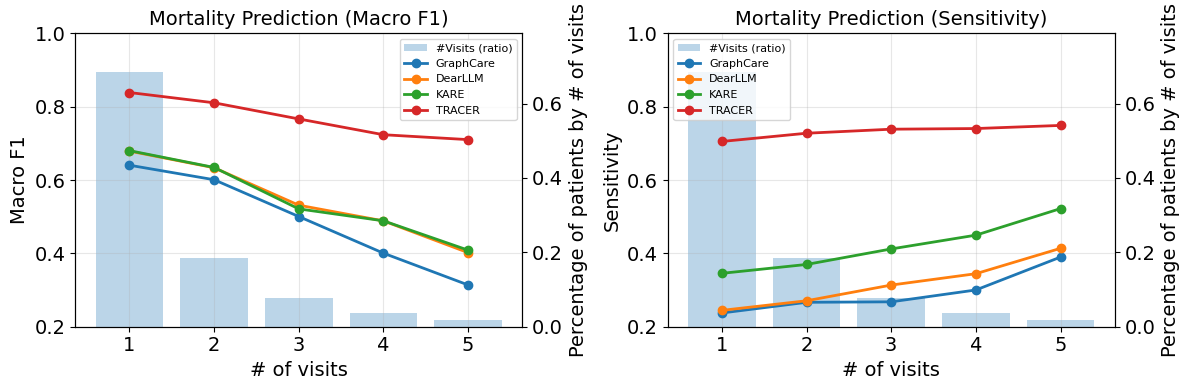}
    \caption{Macro F1 and sensitivity across visit sequence lengths for the mortality prediction task, with the percentage of patients by number of visits.
    }
    \label{fig:intro1}
\end{figure}

\section{Introduction} 
Clinical risk prediction aims to estimate the clinical risk of adverse outcomes, including mortality \cite{intro1, intro2} and readmission \cite{intro3, intro4}, and can support early interventions that help prevent these critical events. To this end, recent methods for clinical prediction rely on Electronic Health Records (EHRs), which contain patients' time-stamped medical histories associated with medical concepts (e.g., diagnoses, procedures, medications), their demographic data, and clinical notes. Building upon EHRs, recent studies have leveraged knowledge graphs (KGs) \cite{GRAM, SeqCare, RetCare} and have retrieved relevant medical information from external literature \cite{intro5} such as PubMed \cite{intro16}. Notably, several approaches \cite{intro8, intro10} construct a patient-specific subgraph of the KG typically associated with medical concepts from a target patient's visit sequence. 

Despite the recent advances in clinical risk prediction, these methods still have several limitations. First, they often struggle to cope with the sparsity and irregularity of patient visit histories in real-world EHRs. Most patients have only a limited number of recorded visits, often separated by heterogeneous and prolonged intervals.

Under such conditions, existing sequential models may fail to capture meaningful temporal dynamics or identify critical turning points in disease progression. 
Although KG-based approaches~\cite{KARE,GraphCARE} attempt to mitigate this issue by incorporating external medical knowledge, they selectively individually extract the medical knowledge related to each of the medical concepts included in a patient's EHR, typically treat all medical concepts equally, ignoring the fact that different diagnoses may have different levels of clinical importance.
As a result, these methods fail to capture severity of concepts and highly plausible disease progressions across visits, thereby being prone to overlook clinically critical “red-flag” events. This limitation is exacerbated by the extreme sparsity of real-world data. As shown in Figure \ref{fig:intro1}, 68.6\% of patients in the \miii dataset have only a single visit, and the vast majority (95.7\%) have fewer than five recorded visits. Under such sparse conditions, state-of-the-art baselines such as GraphCare, DearLLM, and KARE exhibit significant performance degradation. For the mortality prediction task over patients with limited history with $\le$ 5 visits, the Macro F1 scores consistently remain below 0.7 and sensitivity fails to exceed 0.5 except KARE for five visits\footnote{Analogous trends are observed in the readmission prediction task, with details provided in Appendix \ref{append:intro_2}. The decrease in Macro F1 and increase in sensitivity across visit-count groups reflect differences in outcome prevalence across subgroups, and the same trend is observed across all compared models.}.
Second, they underutilize the rich and nuanced information contained in clinical notes. Clinical narratives often encode high-value signals such as physicians’ assessments, suspected complications, treatment responses, and subtle changes in patient condition. However, prior methods either discard these textual sources altogether or employ only a limited set of extracted medical concepts as the patient features. Such simplifications strip away clinically salient context and narrative evidence, which can be crucial for identifying early warning signs and for explaining adverse clinical outcomes.

These limitations provide important insights into three potential solutions: (i) augmenting a patient’s sparse visit sequence with the connected knowledge highly associated with that sequence from medical KGs and the literature, to reveal underlying health progression of her/his health status; and (ii) leveraging the nuanced textual context in clinical notes to capture subjective patients' complains and reactions often underexplored by the existing methods; and
(iii) relying solely on a patient's own history often fails to leverage the clinical experience embedded in historically similar patient cohorts, particularly when personal data is scarce.

Guided by these insights, we propose TRACER, a new retrieval-augmented generation (RAG) framework for trajectory-aware and clinically grounded clinical risk prediction, which explicitly accounts for disease severity, personal narrative evidence, and cohort-level experience. First, to address sparse visit histories and the uniform treatment of diagnoses, TRACER constructs a novel medical knowledge graph called SMKG, grounded on the severity by assigning clinically meaningful  scores to diagnoses based on curated medical literature and clinical guidelines. Severity grounding enables the model to distinguish clinically critical diagnoses and red-flag events from less consequential conditions. Given a patient’s longitudinal visit sequence, TRACER retrieves severity-aware medical trajectories which are sequential paths connecting concepts across adjacent visits that capture clinically plausible disease progressions even under sparse observations. Moreover, to move beyond reliance on an individual patient’s history, TRACER retrieves clinically similar patient cases based on recent visit patterns and trajectories, allowing an LLM to leverage outcome patterns and progression dynamics observed in historically similar cohorts. Second, to fully exploit textual evidence, TRACER enriches each retrieved trajectory with salient evidence from clinical notes, capturing nuanced signals such as clinician assessments, suspected complications, treatment responses, and patient-reported symptoms that are often omitted or oversimplified in structured EHR codes.
By jointly integrating severity-aware KG trajectories, clinical textual evidence, and similar patient retrieval in the RAG framework, TRACER not only improves predictive performance under real-world EHR sparsity but also provides path-level, clinically interpretable explanations.

Our main contributions are summarized as follows:
\begin{itemize}

    \item We propose the severity-grounded medical knowledge graph (SMKG) by assigning fine-grained and diagnosis-level severity scores derived from medical literature.
    \item We address the data sparsity problem by augmenting (a) the target patient's sparse data with patient-specific and severity-aware progression paths across visits from SMKG and (b) clinically similar patient cases, both of which enables the LLM reasoning based on nuanced patient status changes under data sparsity.

    \item We propose to leverage salient and nuanced textual evidence in clinical notes, enabling robust modeling of red-flag events and personal experiences.

    \item We demonstrate consistent and substantial performance gains on the MIMIC-III and MIMIC-IV datasets for two tasks, i.e., mortality prediction and readmission prediction, together with path-level and clinically grounded explanations that enhance interpretability and trust. 
\end{itemize}

\section{Related Work}
Early methods \cite{GRU, Transformer, TCN, Deepr, RETAIN, HiTANet, AdaCare, ConCare, StageNet} for clinical risk prediction primarily focus on modeling the temporal dynamics of patient visit sequence without external knowledge. In contrast, recent approaches focus on two directions: (i) constructing medical KGs from EHR data and extracting subgraphs from external KGs based on patient visit information, (ii) retrieving textual information from external knowledge sources to complement EHR data for clinical risk prediction.

\noindent\textbf{Graph-based Clinical Risk Prediction}. To capture the complex interactions among patient statuses and the relationships between diagnoses and treatments, several methods utilize KGs. GRAM \cite{GRAM} and KAME \cite{KAME} construct medical KGs with attention to enrich patient embeddings. GRASP \cite{GRASP} clusters similar patients into groups and models the relationships between group centroids as a graph. KerPrint \cite{KerPrint} builds time-aware personalized KG to model disease progression. GraphCare \cite{GraphCARE} builds personalized KGs by integrating each patient’s diagnoses, procedures, and medications. However, they extract only 1-hop neighbors of patient-relevant medical concepts in the KGs, and ignore the severity of the medical concepts, thus being unable to capture high-risk events and overlook hidden clinical progressions between visits.

\noindent\textbf{Text-based Clinical Risk Prediction}. Previous text-based methods \cite{MedRetriever, TAPER} retrieve clinically relevant information from structured EHR data. Recently, LLMs have been applied to healthcare tasks due to their advanced reasoning capabilities. KARE~\cite{KARE} integrates EHRs with external knowledge via tailored prompts to generate prediction evidence. DearLLM~\cite{DearLLM} quantifies associations between medical concepts by measuring how likely an LLM is to predict one concept given another, using patient-specific prompts. CPLLM~\cite{CPLLM} fine-tunes LLMs on linearized EHR data. Additionally, several approaches leverage LLMs for clinical text comprehension, such as retrieving and summarizing relevant documents \cite{RAM-EHR} or extracting medical concepts from clinical notes \cite{EMERGE}. However, these methods either neglect clinical notes or fail to fully utilize their rich textual information. They also overlook the severity of diagnoses, resulting in missed nuanced clinical knowledge. 

\section{Preliminary}

\subsection{Electronic Health Records}


A medical concept $c\in C$ refers to a diagnosis, a procedure, or a medication, where $C$ is the set of all medical concepts, e.g., hypertension (diagnosis), appendectomy (procedure), and metformin (medication).
Electronic Health Records (EHRs) comprise visit records, demographic data, and clinical textual records of patients. For each patient, her/his visit records are represented as the temporally ordered visit sequence $[x_1,x_2,\ldots,x_t]$, where each visit $x_i\subseteq C$ consists of the diagnoses, procedures, and medications provided at the patient's $i$-th visit. The demographic $d_i$  contains the patient's attributes available at the $i$-th visit, such as ethnicity, age and gender. We denote a set of clinical notes for the patient at visit $i$ as $\mathcal{R}_i=\{r_{i,1},\ldots,r_{i,m_i}\}$, where each clinical note represent a progress note, discharge summary, or physician note, and $m_i$ denotes the number of clinical notes in $\mathcal{R}_i$. The patient's all clinical textual records are represented as $\mathcal{R}=\{\mathcal{R}_1,\ldots,\mathcal{R}_{t}\}$. More detailed information of EHRs is provided in Appendix~\ref{dataset_stats}.

\subsection{Problem Formulation}
For clinical risk prediction tasks such as mortality prediction and readmission prediction, our goal is to predict specific clinical outcomes based on a patient's EHRs up to that patient's latest visit at $T$. The patient's EHRs are represented as her/his historical visit sequence $[x_1,\ldots,x_T]$, demographic $d_{T}$, and the set $\mathcal{R}$ of clinical notes. The predicted outcomes are defined as follows.

\noindent \textbf{Mortality Prediction.}
Mortality prediction aims to predict a patient's survival status $y\in\{0,1\}$, where 0 indicates the patient will be alive, and 1 indicates the patient will be deceased.

\noindent \textbf{Readmission Prediction.}
Readmission prediction aims to determine a binary outcome $y\in\{0,1\}$ that denotes whether the patient will be readmitted within $\sigma$ days after discharge from the latest historical visit $x_{T}$.  Following previous work~\cite{GraphCARE,KARE}, we set $\sigma=15$ days.


\begin{figure}[t]
    \centering
    \includegraphics[width=1\columnwidth]{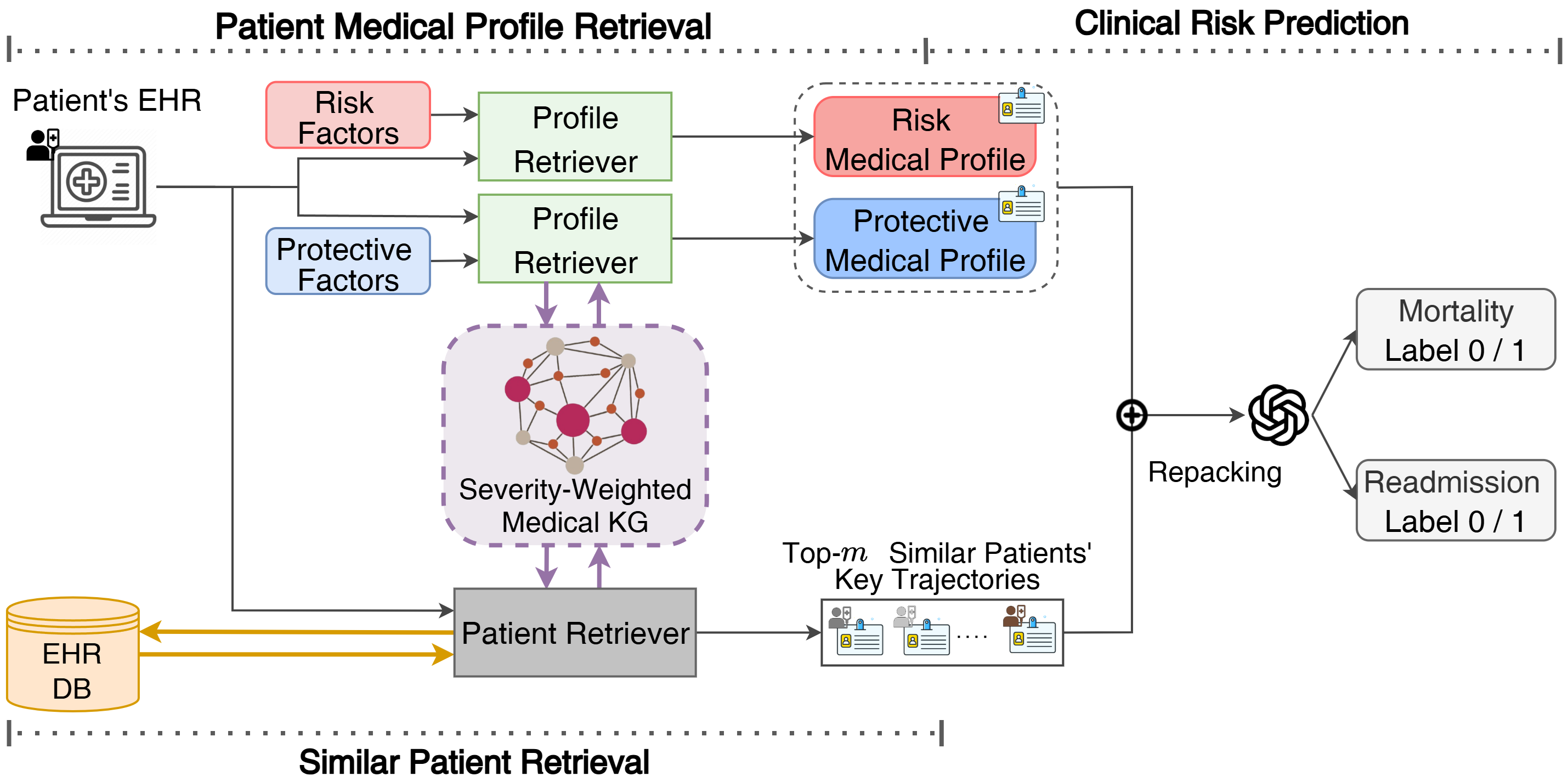}
    \caption{Overview of TRACER
    }
    \label{fig:rc1}
\end{figure}

\section{Methodology}
\subsection{Overview of the TRACER Framework}

Figure \ref{fig:rc1} demonstrates the TRACER framework that consists of three stages: (1) patient medical profile (PMP) retrieval, (2) similar patient retrieval, and (3) clinical risk prediction. 
In stage (1), given a target patient’s EHR and our medical KG, TRACER retrieves the patient's medical profile (PMP), which consists of compact patient-relevant trajectories, relevant passages from the patient's clinical notes, and demographics. In stage (2), we retrieve the trajectories and demographic attributes of top-$m$ patients most similar to the target patient. In stage (3), the LLM generates binary predictions for mortality or readmission, based on the patient's medical profiles, and the top-$m$ similar patients' context. Next, we will introduce our medical KG construction method and describe each of the three main stages.

\subsection{Severity-weighted Medical KG Construction}
\begin{figure}[t]
    \centering
    \includegraphics[width=1\columnwidth]{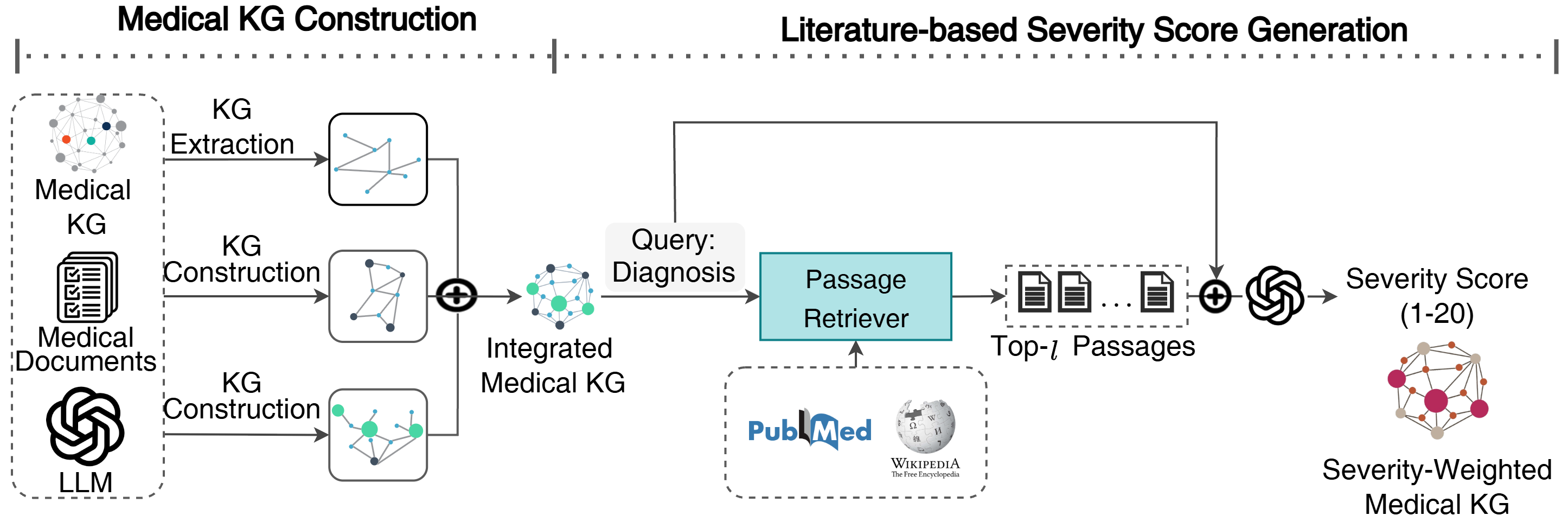}
    \caption{Construction of the severity-weighted medical knowledge graph (SMKG)}
    \label{fig:rc2}
\end{figure}

Despite the importance of considering the severity of diagnoses, existing studies utilizing medical KGs~\cite{ConCare, GRASP, RETAIN} treat all diagnoses equally without reflecting differences in their severity. To address this limitation, we propose a retrieval-augmented severity generation method that quantifies the clinical severity of diagnostic codes, as shown in Figure \ref{fig:rc2}.

\begin{figure*}[t]
    \centering
    \includegraphics[width=1\textwidth]{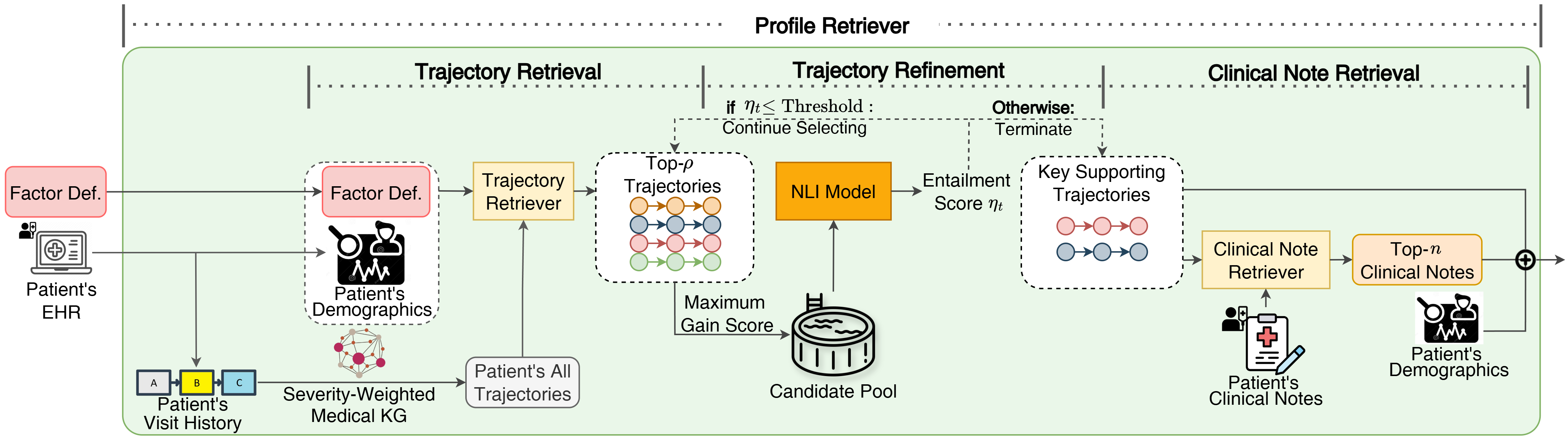}
    \caption{Architecture of the profile retriever}
    \label{fig:rc3}
\end{figure*}

First, we construct a unified medical KG by integrating three data sources: the existing medical KG such as UMLS~\cite{UMLS}, medical documents in PubMed~\cite{intro16}, and LLM-generated triples to capture comprehensive relationships between medical concepts, which have been utilized by KARE~\cite{KARE}. Subsequently, for each diagnostic concept in the unified medical KG, the passage retriever searches all abstracts in PubMed and all summaries in Wikipedia, respectively, for top-$l$ passages closely related to that diagnosis. The LLM takes the concatenation of the retrieved passages as input and generates a severity score for that diagnosis. The score ranges from 1 to 20, where higher values indicate greater clinical severity\footnote{This fine-grained severity scale 1--20 enables the model to distinguish subtle yet clinically meaningful differences among diagnoses, preventing high-risk conditions from being diluted by low-impact or incidental events~\cite{scoring}.}. 
We then define the severity-weighted medical KG (SMKG) as a KG that shares the same node and edge sets as the unified medical KG, where each diagnosis node is assigned the severity score of that diagnosis, and the other types of nodes are assigned a weight of 0\footnote{Other medical concepts such as medications or procedures are excluded from severity scoring, as clinical severity is typically defined for disease conditions rather than interventions or treatments.}.

In subsequent stages, these severity scores enable a more granular capture of a patient’s clinical progression patterns by quantitatively reflecting the clinical severity of each diagnosis, in contrast to existing approaches.  
We provide an analysis of scoring granularity and the score generation prompts in Appendix \ref{appendix_scoring} and Appendix \ref{appendix_Prompt_Severity}, respectively. Additionally, we further validate the clinical relevance of the proposed severity scores by analyzing their correlation with mortality outcomes in Appendix~\ref{appendix_correlation}. 

This LLM-based severity scoring is motivated by recent work showing that LLMs can perform clinically meaningful risk stratification and severity-related assessment~\cite{JMIRRiskStratification, LLMPerioperativeRisk}. Since no universal diagnosis-level severity scale exists across heterogeneous diseases, we additionally assessed the generated scores using a four-level disease-severity framework~\cite{Lazarin2014Severity}. The mean absolute deviation was 0.06 across three runs, corresponding to a 1.6\% error. This indicates that the generated severity scores are reasonably aligned with coarse clinical severity categories. We use severity not as a standalone prognostic label, but as a retrieval prior that is further contextualized by patient-specific trajectories and clinical notes.

\subsection{Patient Medical Profile Retrieval}\label{sec:profile_retrieval}
Most patients' EHR data are sparse. Their visit records typically contain one to five visits, and they are often spaced far apart chronologically. As shown in Figure \ref{fig:rc3}, we address this issue by augmenting the target patient's sparse EHR with her/his structured and unstructured medical history which is highly relevant to the patient and our tasks, referred to as the patient medical profile (PMP). PMP includes (1) the task-relevant progression of the patient's diseases and her/his responses to treatments, (2) the task-relevant descriptions of her/his health status, and (3) the demographic of the patient. For each patient, we retrieve two types of medical profiles focusing on two complimentary clinical factors. The risk factors are associated with adverse outcomes, such as progressive organ dysfunction and high-risk comorbidities; the protective factors are indicative of clinical stability or improvement, such as recent treatment response and stable vital signs.

\subsubsection{Trajectory Retrieval}
We hypothesize that the evolution of a patient's health status can be modeled as a path through the underlying knowledge graph. Formally, we define a trajectory $p$ as a path in the SMKG that connects medical concepts across consecutive visits, following temporal order. 
For every pair of adjacent visits $x_{t-1}$ and $x_t$ in the visit sequence where $t\in\{2,\ldots,T\}$, we consider all cross-visit concept pairs $(c_i,c_j)\in x_{t-1}\times x_t$. For each pair, we extract all shortest paths in the SMKG up to a maximum length of $h=3$ hops to ensure both clinical relevance and computational tractability.
Let $\mathcal{P}_t$ denote the set of all such visit-to-visit paths between $x_{t-1}$ and $x_t$. The complete set of candidate trajectories is then formed by the Cartesian product $\mathcal{P}_2\times\cdots\times\mathcal{P}_{T}$, where each candidate trajectory corresponds to a unique sequential concatenation of paths selected from $\mathcal{P}_2,\cdots,\mathcal{P}_{T}$
\footnote{Single-visit patients are excluded from the cohort; for each retained patient, the final visit serves as the prediction target, while preceding visits constitute the historical context.}. We compute the embedding $\vect{e}_p$ of each candidate trajectory using the pretrained biomedical encoder BioClinicalBERT \cite{BioClinicalBERT}.

We preserve these path-level trajectories rather than merging them into a single unified patient graph; this design allows them to be independently ranked, diversified, repacked, and directly verbalized as explicit, interpretable evidence for the LLM.

To capture the full spectrum of a patient's medical profile, we perform a dual-retrieval process targeting both risk-inducing and protective trajectories. We define risk factors as clinical evidence associated with adverse outcomes, such as progressive organ dysfunction, severe comorbidities, or deteriorating physiological status. In contrast, protective factors refer to evidence indicating clinical stability or recovery, such as treatment response, stable vital signs, or improving symptoms~\cite{KraemerRiskTerms, HFimpEF}. 

Based on these definitions, we construct two distinct textual queries: a risk query $q_r$ and a protective query $q_o$. Each query is composed of: (1) the specific prediction task description and (2) the patient’s demographics $d_{T}$. For every candidate trajectory $p$, we compute two relevance scores $\psi_r$ and $\psi_o$ for risk and protective queries, respectively:

\begin{equation*}
    \psi_r(q_r, p) = \alpha \cos(\vect{e}_{q_r}, \vect{e}_p) + (1 - \alpha) s(p)
\end{equation*}
\begin{equation*}
    \psi_o(q_o, p) = \beta \cos(\vect{e}_{q_o}, \vect{e}_p) + (1 - \beta)(1 - s(p))
\end{equation*}

\noindent where $\vect{e}_{q_r}$ and $\vect{e}_{q_o}$ denote the embeddings of the respective query, and the severity $s(p)$ of trajectory $p$ is calculated as the average of the normalized severity score (ranging $[0,1]$) of every diagnosis node contained in $p$. Risk and protective aspects are inherently complementary: an increase in one severity implies a decrease in the other. By using $1 - s(p)$ for $\psi_o$, we maintain interpretive consistency and joint expressiveness. Hyperparameters $\alpha, \beta 
\in [0,1]$ weigh the relative importance of the semantic relevance to the queries and the severity of the trajectory. Finally, we retrieve the top $\rho$\% of trajectories for each aspect based on these scores. This linear score is not used as a standalone risk estimator; rather, it serves as an interpretable retrieval prior that prioritizes trajectories that are both clinically severe and relevant to the target prediction task. The prompt for trajectory retrieval is provided in Appendix \ref{appendix_prompt_reconstruct}.

\subsubsection{Trajectory Refinement}
While coarse first-pass retrieval in the previous stage provides query-relevant results, it often yields candidates that lack precise relevance and exhibit redundancy. Such redundancy may
not only reduce the diversity of retrieved evidence but also obscure clinically meaningful progression patterns that are critical for downstream prediction and interpretation. In the trajectory refinement stage, we thus rerank the retrieved trajectories, and filter out the redundant trajectories among them to achieve both the query-trajectory relevance and diversity. Specifically, we initialize an empty pool $S_0$ of trajectories. Next, we iteratively refine them based on maximum marginal relevance (MMR) to balance between the relevance of the trajectories to the aforementioned risk (spec. protective query) and their novelty with respect to the already selected set $S_{i-1}$. At each iteration $i$, we compute the below gain score for each of the retrieved trajectories not included in $S_{i-1}$, and add the trajectory $p^*$ with the highest gain score to the pool, i.e., $S_i = S_{i-1} \cup \{p^*\}$.
\begin{equation*}
    g(p_i) = \lambda_{MMR}\cos(\vect{e}_q, \vect{e}_{p_i}) - (1 - \lambda_{MMR}) \max_{j \in S_{i-1}} \cos(\vect{e}_{p_i}, \vect{e}_{p_j})
\end{equation*}

\noindent where hyperparameter $\lambda_{MMR}$ is a weighting factor. Beyond semantic similarity, we further assess whether the selected trajectory set collectively entails the clinical conditions and diagnoses closely associated with the target patient. We calculate the entailment score $\eta_i$ of $S_i$ by using the pretrained NLI model SimCSE-Roberta \cite{SimCSE}:
\begin{equation*}
    \eta_i = NLI(\{ p \vert p \in S_i \}, q)
\end{equation*}

\noindent where $NLI(\cdot, \cdot)$ takes as input the text for all trajectories $p \in S_i$  and the query $q$, i.e., either $q_r$ or $q_o$. The iteration terminates when either $\eta_i \geq \lambda_{max}$ or $\eta_i- \eta_{i-1} \leq \lambda_{min}$ is satisfied where thresholds $\lambda_{max}$ and $\lambda_{min}$ are hyperparameters. 
This combination of MMR and NLI-based stopping preserves relevant, diverse, and sufficiently supportive trajectories, aligned with prior retrieval-augmented evidence selection paradigms \cite{BIDER}. We denote the trajectories remaining in the candidate pool after termination as Key Supporting Trajectories (KSTs).
Finding the KSTs can be achieved solely by designing the prompt detailed in Appendix \ref{appendix_prompt_reconstruct} and adjusting the thresholds, without requiring additional model training. The KSTs compose context provided to the LLM that performs the prediction tasks. 

\begin{table*}[t]
\centering
\caption{Overall performance of two tasks on MIMIC-III. The numbers in parentheses indicate the proportion of patients with positive labels, i.e., 1s, for each task. The best and second best results are denoted in bold and underlined, respectively.}
\label{tab:main1}
\renewcommand{\arraystretch}{0.80} 
\resizebox{\textwidth}{!}{%
{\fontsize{6}{7}\selectfont
\begin{tabular}{lcccccccc}
\toprule
\multirow{2}{*}{\textbf{Models}} & \multicolumn{4}{c}{\textbf{Mortality Prediction (10.83\%)}} & \multicolumn{4}{c}{\textbf{Readmission Prediction (21.6\%)}} \\
\cmidrule(lr){2-5} \cmidrule(lr){6-9}
& Macro F1 & Sensitivity & Specificity & Accuracy & Macro F1 & Sensitivity & Specificity & Accuracy \\
\midrule
GRASP       & 0.5176 & 0.0495 & \textbf{0.9963} & 0.8938 & 0.5262 & 0.1666 & 0.8897 & 0.7335 \\
StageNet   & 0.5571 & 0.1881 & 0.9189 & 0.8390 & 0.5275 & 0.1666 & \underline{0.8923} & 0.7355 \\
MedRetriever & 0.5622 & 0.2079 & 0.9122 & 0.8358 & 0.5746 & 0.6524 & 0.6168 & 0.6245 \\
GraphCare   & 0.5963 & 0.2552 & 0.9577 & 0.9229 & 0.6364 & 0.6999 & 0.5964 & 0.6779 \\
RAM-EHR    & 0.6165 & 0.2277 & \underline{0.9627} & 0.8830 & 0.5678 & 0.7143 & 0.5761 & 0.6056 \\
DearLLM    & 0.6374 & 0.2673 & 0.9615 & 0.8863 & 0.6390 & 0.7095 & 0.6194 & 0.7343 \\
KARE       & \underline{0.6375} & \underline{0.3711} & 0.9593 & \underline{0.9267} & \underline{0.7271} & \underline{0.7501} & 0.7112 & \underline{0.7451} \\
TRACER     & \textbf{0.8197} & \textbf{0.7128} & 0.9531 & \textbf{0.9275} & \textbf{0.8709} & \textbf{0.7809} & \textbf{0.9501} & \textbf{0.9135} \\
\bottomrule
\end{tabular}
}%
}
\end{table*}

\begin{table*}[t]
\centering
\caption{Overall performance of two tasks on \miv. The numbers in parentheses indicate the proportion of patients with positive labels, i.e., 1s, for each task. The best and second best results are denoted in bold and underlined, respectively.}
\label{tab:main2}
\renewcommand{\arraystretch}{0.80} 
\resizebox{\textwidth}{!}{%
{\fontsize{6}{7}\selectfont
\begin{tabular}{lcccccccc}
\toprule
\multirow{2}{*}{\textbf{Models}} & \multicolumn{4}{c}{\textbf{Mortality Prediction (38.13\%)}} & \multicolumn{4}{c}{\textbf{Readmission Prediction (48.84\%)}} \\
\cmidrule(lr){2-5} \cmidrule(lr){6-9}
& Macro F1 & Sensitivity & Specificity & Accuracy & Macro F1 & Sensitivity & Specificity & Accuracy \\
\midrule
GRASP       & 0.7514 & 0.5063 & \underline{0.9688} & 0.7925 & 0.5876 & 0.5980 & 0.5766 & 0.5881 \\
StageNet    & 0.7479 & 0.5494 & 0.9236 & 0.7809 & 0.6108 & 0.6594 & 0.5652 & 0.6112 \\
MedRetriever & 0.7727 & 0.5797 & 0.9392 & 0.8021 & 0.6237 & 0.6931 & 0.5595 & 0.6247 \\
GraphCare   & 0.8356 & 0.6327 & 0.9491 & 0.8782 & 0.6713 & 0.6831 & 0.6219 & 0.6493 \\
RAM-EHR     & 0.8223 & 0.6303 & 0.9579 & 0.8460 & 0.6374 & 0.6495 & 0.6257 & 0.6373 \\
DearLLM     & 0.8130 & 0.6456 & 0.9516 & 0.8349 & 0.6870 & 0.7485 & 0.6295 & 0.6876 \\
KARE       & \underline{0.8698} & \underline{0.7498} & \textbf{0.9771} & \underline{0.8893} & \underline{0.7444} & \underline{0.8582} & \underline{0.6469} & \underline{0.7256} \\
TRACER     & \textbf{0.8830} & \textbf{0.7722} & \underline{0.9688} & \textbf{0.8938} & \textbf{0.7942} & \textbf{0.8752} & \textbf{0.7183} & \textbf{0.7949} \\
\bottomrule
\end{tabular}
}%
}
\end{table*}

\subsubsection{Clinical Note Retrieval}
Patient clinical notes are unstructured documents that comprehensively record various information including physicians’ subjective assessments, detailed laboratory measurements, symptom descriptions, and the rationale behind prescriptions. Such unstructured narratives often contain early warning signs and nuanced clinical cues that are not explicitly captured by structured EHR codes. However, utilizing all clinical notes without filtering may introduce irrelevant information, which can degrade the performance of predictive models, e.g., LLMs.

To address this issue, we retrieve relevant passages from the set $\mathcal{R}$ of the patient's historical clinical records. Recall that $\mathcal{R}=\{\mathcal{R}_1,\ldots,\mathcal{R}_{T}\}$ contains all clinical textual records associated with the patient's historical visits. We construct the passage pool $\mathcal{B}$ by splitting every clinical note in $\mathcal{R}$ into passages. For each of risk query $q_r$ and protective query $q_o$, we embed the query and each passage $b\in\mathcal{B}$ with a pretrained encoder, compute cosine similarities between the query and passage embeddings, and select the top-$n$ passages most relevant to the query. The retrieved passages are then used as note-level evidence in the patient medical profile.


\subsection{Retrieval-augmented Clinical Risk Prediction}
Patients’ EHRs typically consist of a small number of visits. However, the existing methods \cite{StageNet, DearLLM} rely solely on a patient’s own records, and embed the entire visit sequence into single vectors, which often introduce irrelevant or outdated information. 
They also \cite{PRISM, ColaCare} do not take advantage of valuable information hidden in their unstructured clinical notes.
To address these limitations, we augment a target patient’s EHR with \textbf{three types of supporting evidence}: (A) the patient’s risk and protective medical profiles, (B) the patient's demographics, and (C) the KSTs and demographics of her/his top-$m$ similar patients. Specifically, the risk and protective medical profiles are produced by the profile retriever described in Section \ref{sec:profile_retrieval}. For evidence (C), we first calculate the Jaccard similarity between the set of medical concepts in the target patient’s recent visits and those from all other patients, and we then select the top‑$m$ similar patients based on the similarity. Next, we concatenate the healthcare task description, and evidence (A), (B) and (C) to form  a single prompt. During this, we perform repacking on the KSTs and clinical notes, respectively, in evidence (A) so that the more relevant KST and clinical note come earlier and are thus located closer to the task description in the prompt.  This repacking strategy allows the LLM to focus on more relevant information \cite{Repacking1, Repacking2}. 
Eventually, the LLM generates a binary outcome $y\in \{0, 1\}$.

By harmonizing (i) retrieval-augmentation with the severity-aware trajectories, the relevant clinical notes, and similar peer cases, (ii) dynamic context repacking, and (iii) a clear step-by-step reasoning process, our framework mitigates the “lost‑in‑the‑middle” problem \cite{Lost-in-the-middle} inherent to long-context LLM inputs. It also addresses both sparsity in time‑series data and redundancy in unstructured notes. This retrieval‑augmented, repacking‑guided reasoning not only improves predictive performance but also generates transparent and interpretable chains of evidence that substantiate each risk estimate, thereby enhancing the clinical trustworthiness of the predictions. The impact of similar patients and the analysis of the repacking strategy are provided in Section \ref{sec:ablation1} and Appendix \ref{ablation2}, respectively.

\begin{table}[t]
\centering
\caption{Ablation study on MIMIC-III. The variants are grouped by major evidence modules. Standard deviations over three runs are below 0.06 for all variants.}
\label{tab:ablation}
\resizebox{\columnwidth}{!}{
\begin{tabular}{llcccc}
\toprule
\textbf{Module} & \textbf{Variant} 
& \multicolumn{2}{c}{\textbf{Mortality}} 
& \multicolumn{2}{c}{\textbf{Readmission}} \\
\cmidrule(lr){3-4} \cmidrule(lr){5-6}
& & Macro F1 & Sensitivity & Macro F1 & Sensitivity \\
\midrule
Full model 
& TRACER 
& \textbf{0.8197} & \textbf{0.7128} 
& \textbf{0.8709} & \textbf{0.7809} \\
\midrule
SMKG 
& w/o Sev. 
& 0.7484 & 0.4950 
& 0.8643 & 0.7571 \\
\midrule
\multirow{3}{*}{Trajectory}
& w/o Risk. 
& 0.7328 & 0.4356 
& 0.8499 & 0.7095 \\
& w/o Prot. 
& 0.7318 & 0.5247 
& 0.8507 & 0.7143 \\
& w/o Traj. 
& 0.6302 & 0.4653 
& 0.8103 & 0.7000 \\
\midrule
Clinical notes
& w/o Clin. 
& 0.7757 & 0.5049 
& 0.7981 & 0.6762 \\
\midrule
Similar patients
& w/o Pat. 
& 0.7527 & 0.5049 
& 0.8050 & 0.7047 \\
\midrule
\multirow{2}{*}{Combined}
& w/o PMP 
& 0.6656 & 0.2475 
& 0.7381 & 0.5285 \\
& w/o (PMP+Pat.) 
& 0.6424 & 0.2173 
& 0.6999 & 0.4952 \\
\bottomrule
\end{tabular}}
\end{table}

\section{Experiments}
\subsection{Experimental Setup}
\textbf{Datasets.} 
We utilize two publicly available datasets, \miii \cite{MIMIC-III} and \miv \cite{MIMIC-IV}. The mortality rates are 10.83\% and 38.13\%, while the readmission rates are 21.6\% and 48.84\% in \miii and \miv, respectively. 
Appendix \ref{dataset_det} details the label definition. 

\noindent\textbf{Evaluation Metrics.} We assess performance using four metrics: Macro F1 score, sensitivity, specificity, and accuracy. A higher value indicates better performance. More details on the metrics are described in Appendix \ref{appendix_metrics}.

\noindent\textbf{Baseline Methods}. We adopt state-of-the-art healthcare prediction models: StageNet \cite{StageNet}, GRASP \cite{GRASP}, MedRetriever \cite{MedRetriever}, RAM-EHR \cite{RAM-EHR}, DearLLM \cite{DearLLM}, GraphCare \cite{GraphCARE}, and KARE \cite{KARE}\footnote{The preprocessing codes of EMERGE \cite{EMERGE} and KerPrint \cite{KerPrint} are not public, and the reported results for these models use incorrect preprocessing, so we do not include their experimental results.}. We further compare TRACER with a recent EHR foundation model, ETHOS, and a prompt-only zero-shot LLM baseline to isolate the effect of retrieval augmentation. We showcase the details of our baseline implementation in Appendix \ref{appendix_baselines}.
 
\noindent \textbf{Implementation Details.} We conduct all experiments on a server equipped with an Intel Xeon(R) Gold 6226R CPU, a single NVIDIA RTX A6000 GPU (48 GB VRAM), and 256 GB of DDR4 RAM. 
For risk and protective trajectory retrieval, $\alpha$ and $\beta$  are set to 0.8 and 0.2 respectively. $\lambda_{MMR}$ in the trajectory refinement stage is set to 0.7, and the thresholds $\lambda_{max}$ and $\lambda_{min}$ for the entailment scores are 0.5 and 0.05 respectively. For clinical notes and similar patients retrieval, we set both $n$ and $m$ to $5$. We leverage OpenAI GPT-4o-mini \cite{GPT-4o-mini} to assign diagnostic severity scores and to perform retrieval-augmented clinical risk prediction. For fair comparison, we use GPT-4o-mini as the consistent backbone LLM for all LLM-based baseline models.


\subsection{Main Results}

\begin{table*}[t]
\centering
\caption{Performance comparison on MIMIC-III across different LLMs for mortality and readmission prediction, along with severity score variability, where lower values indicate more consistent severity score generation.}
\label{tab:consistency_var}
\resizebox{\textwidth}{!}{%
\begin{tabular}{lccccccccc}
\toprule
\textbf{LLM} &
\multicolumn{4}{c}{\textbf{Mortality Prediction}} &
\multicolumn{4}{c}{\textbf{Readmission Prediction}} &
\textbf{Severity S.D.} \\
\cmidrule(lr){2-5}\cmidrule(lr){6-9}
 & \textbf{Accuracy} & \textbf{Macro F1} & \textbf{Sensitivity} & \textbf{Specificity}
 & \textbf{Accuracy} & \textbf{Macro F1} & \textbf{Sensitivity} & \textbf{Specificity}
 &  \\
\midrule
GPT-4o-mini & 0.9275 & 0.8197 & 0.7128 & 0.9531 & 0.9135 & 0.8709 & 0.7809 & 0.9501 & 2.02 \\
GPT-5-nano  & 0.9356 & 0.8289 & 0.7311 & 0.9608 & 0.9222 & 0.8805 & 0.7963 & 0.9598 & 1.59 \\
GPT-o3      & \textbf{0.9460} & \textbf{0.8447} & \textbf{0.7562} & \textbf{0.9701} & \textbf{0.9471} & \textbf{0.8916} & \textbf{0.8186} & \textbf{0.9701} & \textbf{1.16} \\
\bottomrule
\end{tabular}%
}
\end{table*}

Table \ref{tab:main1} compares the performance of mortality prediction and readmission prediction for TRACER and the latest baselines on MIMIC-III. TRACER outperforms all competitors across all metrics on both tasks. For the mortality prediction task, sensitivity and Macro F1 score improve by 28.5\% and 19.7\%, respectively, compared to the state-of-the-art method, i.e., KARE. We speculate that this gap may be due to the patient-specific trajectories sampled solely from our severity-weighted medical KG.
Table \ref{tab:main2} shows the performance of the compared methods on \miv. For the mortality prediction task, TRACER achieves the state of the art performance across all metrics except specificity, though it trails the best competitor by only 0.8\%. In contrast, for the readmission prediction task, TRACER surpasses the competitors on specificity by at least 15.5\%. These results demonstrate that TRACER is highly competitive for both tasks. In addition, the observed performance gains of TRACER are consistent across different model categories. Specifically, models relying solely on EHR sequences (e.g., GRASP, StageNet, RAM-EHR) are consistently outperformed by approaches that incorporate external knowledge or additional features, such as medical KGs (e.g., KARE) or retrieved medical concepts (e.g., RAM-EHR, MedRetriever). 
TRACER further advances this trend by leveraging severity-weighted KG trajectories and clinical text retrieval, resulting in a clear and monotonic improvement from these existing methods.
Importantly, the improvement achieved by TRACER substantially exceeds the typical performance gains observed in prior work. Across existing baselines, the average incremental improvement in Macro F1 over preceding methods is approximately 2.0 points, whereas TRACER improves Macro F1 by 18.2 points over the strongest prior model (KARE). This indicates that TRACER delivers a substantive methodological advance beyond incremental model refinements. All reported results are averaged over three runs; standard deviations are below 0.06 across the main variants, indicating stable trends. More detailed analysis on mortality and readmission prediction performances are included in Appendix \ref{append:main_results}.

\subsection{Ablation Study}\label{sec:ablation1}
We conduct an ablation study to examine the contribution of each component in TRACER. To clarify attribution in the multi-component pipeline, we organize the variants around core modules: SMKG, trajectory retrieval, clinical note retrieval, and similar-patient retrieval. We also evaluate the variants with combined modules removed to assess whether the overall performance is driven by a single component or by complementary integration. All ablation results are averaged over three runs, and the standard deviation of each variant is below 0.06, indicating stable trends across runs.

The variant ``w/o Sev.'' samples trajectories from the medical KG without severity scores; ``w/o Risk.'', ``w/o Prot.'', and ``w/o Traj.'' denote the variants in which the LLM performs reasoning based on all the proposed supporting evidence except risk trajectories, except protective trajectories, and except both trajectories, respectively; ``w/o Pat.'' denotes the variant that builds the proposed supporting evidence except only the trajectories and demographics of the top-$m$ similar patients, and ``w/o Clin.'' represents the variant with all the supporting evidence except top-$n$ clinical notes. We also evaluate a variant with the patient medical profile (w/o PMP) removed or that with both PMP and top-$m$ patients removed (w/o (PMP+Pat.)).

Since MIMIC-III and MIMIC-IV are highly imbalanced, sensitivity is the most important metric~\cite{KARE} because failing to identify true positive cases such as actual mortality or readmission can have critical clinical consequences.

Table~\ref{tab:ablation} shows that the three core modules contribute complementary gains. Removing severity from SMKG reduces mortality sensitivity from 0.7128 to 0.4950, confirming the role of severity grounding. Trajectory retrieval is the most influential component for mortality prediction, as removing all trajectories reduces Macro F1 from 0.8197 to 0.6302. In particular, removing risk trajectories causes the largest sensitivity drop, indicating that severity-aware disease progression is critical for detecting mortality cases.
In contrast, readmission prediction is more sensitive to clinical notes and similar-patient evidence. Removing clinical notes decreases Macro F1 from 0.8709 to 0.7981 and sensitivity from 0.7809 to 0.6762, suggesting that unstructured notes capture short-term deterioration and discharge-related cues that are not fully represented by structured codes. Removing similar patients causes a clear performance drop, indicating that cohort-level context provides useful complementary evidence. Finally, the combined ablations show that TRACER is not driven by a single component, but by the coordinated use of SMKG, trajectory retrieval, clinical notes, and similar-patient evidence.

\section{Analysis}
\subsection{Consistency of Severity Score Generation}
To clarify the consistency of severity score generation, we generate each disease-level severity score three times using different LLM backbones and evaluate the resulting end-to-end pipeline on MIMIC-III for both mortality and readmission prediction. For each disease, we also calculate the standard deviation, dubbed Severity S.D., from the mean of the three generated scores. Lower values indicate more consistent severity estimation across LLM backbones.

Table~\ref{tab:consistency_var} summarizes end-to-end prediction performance and severity-score consistency across different LLM backbones. Severity S.D. decreases from 2.02 with GPT-4o-mini to 1.59 with GPT-5-nano and 1.16 with GPT-o3, indicating that larger models generate more consistent severity scores for the same disease. A similar trend is observed in both mortality and readmission prediction performance, where larger models achieve higher accuracy. In particular, GPT-o3 more effectively exploits severity-aware trajectories and retrieved clinical notes, resulting in over a 3\% improvement in Macro F1 for mortality prediction compared to GPT-4o-mini. Importantly, severity variability remains low even for GPT-4o-mini, demonstrating that the proposed severity scoring procedure is robust and not overly sensitive to the choice of LLM.

\subsection{Effectiveness of SMKG}
\begin{table}[t]
\centering
\caption{Performance comparison of MedGraphRAG when replacing the UMLS graph with the proposed SMKG across multiple medical QA benchmarks.}
\label{tab:smkg_adjustment}
\resizebox{\columnwidth}{!}{%
\begin{tabular}{lccc}
\toprule
\textbf{Methods} & \textbf{MedMCQA} & \textbf{PubMedQA} & \textbf{MedQA} \\
\midrule
MedGraphRAG - UMLS & 81.5 & 83.3 & 91.3 \\
MedGraphRAG - SMKG & \textbf{84.2} & \textbf{86.5} & \textbf{94.9} \\
\bottomrule
\end{tabular}%
}
\end{table}


To evaluate the effectiveness of SMKG, we replace the UMLS-based medical knowledge graph used in MedGraphRAG \cite{MedicalGraphRAG} with the proposed SMKG, and analyze the resulting performance changes. Experiments are conducted on three representative medical question answering benchmarks: MedQA \cite{MedQA}, MedMCQA \cite{MedMCQA}, and PubMedQA \cite{PubMedQA}. Table \ref{tab:smkg_adjustment} show that substituting UMLS with SMKG consistently improves performance across all benchmarks. In particular, on the MedQA dataset, the accuracy increases by 3.94 percentage points, representing the most substantial gain among the evaluated datasets. These findings indicate that, unlike conventional medical knowledge graphs that treat all disease concepts with equal importance, assigning severity scores to disease nodes enables the model to more effectively distinguish relative clinical risk and importance among diseases. Overall, SMKG enhances medical question answering performance by explicitly encoding disease severity into the graph structure, allowing the model to prioritize clinically higher-risk conditions and retrieve more relevant evidence during reasoning.

\subsection{Inference efficiency}

\begin{table}[t]
\centering
\caption{Comparison of inference efficiency between \textbf{TRACER} and \textbf{KARE} on the MIMIC-III for mortality prediction task.}
\label{tab:efficiency}
\resizebox{\columnwidth}{!}{%
\begin{tabular}{lcccc}
\toprule
\textbf{Model} 
& \begin{tabular}[c]{@{}c@{}}\textbf{Inference}\\\textbf{Time (s)}\end{tabular}
& \begin{tabular}[c]{@{}c@{}}\textbf{Patient-specific Context}\\\textbf{Retrieval Time (s)}\end{tabular}
& \textbf{Macro F1} 
& \textbf{Sensitivity} \\
\midrule
KARE   & 117 & 92 & 0.6375 & 0.3711 \\
TRACER & 88  & 60 & \textbf{0.8284} & \textbf{0.6831} \\
\bottomrule
\end{tabular}%
}
\end{table}

Table \ref{tab:efficiency} compares the inference time and accuracy of  TRACER and KARE, i.e., the best competitor in terms of overall prediction performance\footnote{For TRACER, inference time is measured as the total duration of the inference stage. For KARE, inference time is defined as the sum of the durations required for positive/negative patient retrieval, reasoning chain construction, and LLM prediction.}. The results show that TRACER is approximately 33\% faster than KARE in terms of inference time, while achieving 29.9\% and 84.1\% higher Macro F1 and sensitivity, respectively. 
Both methods perform patient-specific context retrieval to support reasoning-based prediction, but differ substantially in how such information is retrieved. KARE performs patient subgraph retrieval by iteratively expanding multi-hop neighborhoods from patient-related medical concepts in its medical knowledge graph, while TRACER retrieves severity-aware, temporally ordered disease trajectories tailored to the target patient from the SMKG.
As shown in Table \ref{tab:efficiency}, the patient subgraph retrieval time for KARE is 92 seconds, while the patient trajectory retrieval for TRACER requires only 60 seconds, making TRACER about 35\% faster. This demonstrates that leveraging trajectories offers clear advantages in both efficiency and predictive accuracy.

\begin{figure}[t]
    \centering
    \includegraphics[width=\columnwidth]{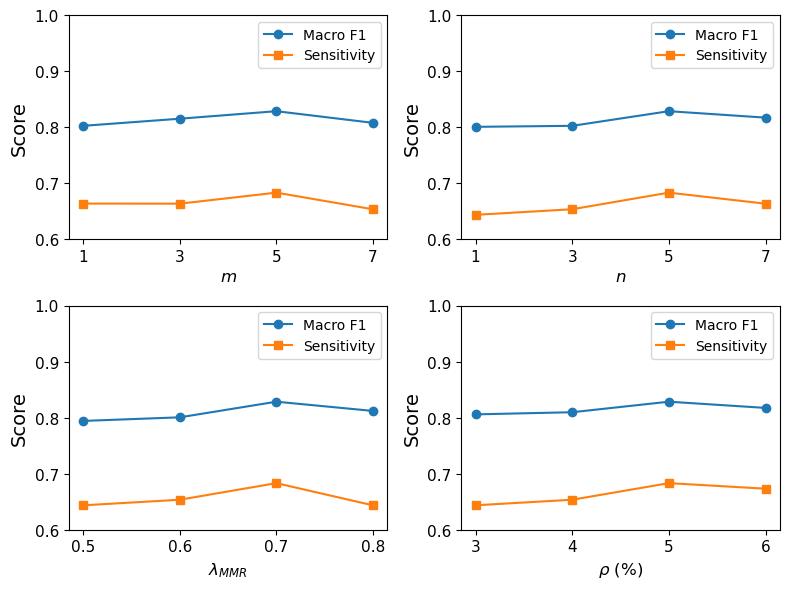}
    \caption{Sensitivity analysis of the number of similar patients, the number of clinical notes, the proportion of trajectories, and the weight for the MMR score on MIMIC-III.}
    \label{fig:hyperparameters}
\end{figure}

\subsection{Hyperparameter Sensitivity}


We evaluate the impact of hyperparameters on mortality prediction performance on MIMIC-III. Figure \ref{fig:hyperparameters} shows the changes in Macro F1 score and sensitivity for: (i) the number $m$ of similar patients, (ii) the number $n$ of clinical notes, (iii) the proportion $\rho$ of trajectories, and (iv) weight factor $\lambda_{MMR}$ for MMR score.
As described in Figure \ref{fig:hyperparameters}, $m$ and $n$ exhibit similar trends, achieving the highest performance when both are set to 5, with Macro F1 score and sensitivity reaching 0.8284 and 0.6831, respectively. When set below 5, the performance decreases slightly, while values above 5 lead to a drop in performance due to excessive information. The weighting factor $\lambda_{MMR}$ for the MMR score achieves the best performance at 0.7, but shows a decline when set above 0.7. In addition, the proportion $\rho$ of trajectories most relevant to each patient and each (risk or protective) factor yields the best results at 5, while performance slightly drops for values less than 5. Further details are presented in Appendix \ref{appendix_hyperparameters}.

\begin{table}[t]
\centering
\caption{Comparison of prediction performance according to the method for selecting similar patients on MIMIC-IV. The best results are denoted in bold.}
\label{tab:similar_patients}
\resizebox{\columnwidth}{!}{%
\begin{tabular}{lcccc}
\toprule
\multirow{2}{*}{\textbf{Methods}} 
  & \multicolumn{2}{c}{\textbf{Mortality Prediction (38.13\%)}} 
  & \multicolumn{2}{c}{\textbf{Readmission Prediction (48.84\%)}} \\
\cmidrule(lr){2-3} \cmidrule(lr){4-5}
 & Macro F1 & Sensitivity & Macro F1 & Sensitivity \\
\midrule
StageNet embedding & 0.8798 & 0.7696 & 0.7907 & 0.8732 \\
Last-visit Jaccard & 0.8787 & 0.7646 & 0.7904 & 0.8712 \\
Recent-visit Jaccard & \textbf{0.8830} & \textbf{0.7722} & \textbf{0.7942} & \textbf{0.8752} \\
\bottomrule
\end{tabular}%
}
\end{table}

\subsection{Similar Patient Retrieval Strategy}

We conduct experiments to compare the performance of mortality and readmission prediction according to the criterion used to retrieve similar patients. As shown in Table~\ref{tab:similar_patients}, we compare three strategies: (i) StageNet embedding, which retrieves similar patients using cosine similarity over the entire visit-sequence embedding; (ii) Last-visit Jaccard, which uses Jaccard similarity over medical concepts appearing only in the last historical visit; and (iii) Recent-visit Jaccard, which uses Jaccard similarity over medical concepts appearing in the most recent two historical visits.

The results show that Recent-visit Jaccard achieves the best performance across both tasks. This supports our design choice of retrieving similar patients based on exact recent overlap rather than broad sequence-level semantic similarity. In clinical risk prediction, semantically related concepts may still imply different prognostic risks, e.g., sepsis and septic shock are closely related but differ substantially in mortality risk. Beyond predictive performance, Recent-visit Jaccard achieves an average retrieval speed per patient that is approximately 19\% faster than the StageNet baseline. Balancing both effectiveness and efficiency, Recent-visit Jaccard emerges as the most practical choice\footnote{Similar-patient evidence is used only as auxiliary context: the final prediction remains primarily grounded in the target patient's own trajectories, clinical notes, and demographics, thereby reducing the risk of shortcut reasoning based only on peer outcomes.}.

\begin{table}[t]
\centering
\caption{Robustness analysis across different LLM backbones on MIMIC-III. The best results are denoted in bold.}
\label{tab:robustness_llm_backbones}
{\fontsize{8}{8}\selectfont
\resizebox{\columnwidth}{!}{%
\begin{tabular}{lcccc}
\toprule
\textbf{Methods} & Macro F1 & Sensitivity & Specificity & Accuracy \\
\midrule
GPT-4o-mini & \textbf{0.8197} & \textbf{0.7128} & 0.9531 & \textbf{0.9275} \\
LLaMA2-8B & 0.7729 & 0.5446 & \textbf{0.9638} & 0.9182 \\
Qwen3.5-2B & 0.7581 & 0.5842 & 0.9434 & 0.9045 \\
\bottomrule
\end{tabular}%
}
}
\end{table}

\subsection{Robustness to LLM Backbones}\label{sec:robustness_llm_backbones}
Table~\ref{tab:robustness_llm_backbones} evaluates the robustness of TRACER across different LLM backbones on MIMIC-III mortality prediction. GPT-4o-mini achieves the best overall performance, while the open-source LLMs also maintain competitive results. In particular, LLaMA2-8B obtains the highest specificity, and Qwen3.5-2B preserves reasonable sensitivity and accuracy despite its smaller scale. These results indicate that TRACER is not specific to GPT-family models, but can remain effective when instantiated with different LLM backbones. We further provide a comparison across cutting-edge clinical risk prediction baselines and robustness analysis across knowledge snapshots in Appendices~\ref{sec:appendix_main_result_baselines} and~\ref{sec:appendix_robustness_snapshots}, respectively.



\subsection{Hop-limit Analyses}
\label{sec:hop_coverage_analysis}


We evaluate the impact of the maximum path length $h$ during trajectory extraction to determine the optimal amount of KG context. As shown in Table~\ref{tab:design_compact}, a restrictive hop limit risks omitting clinically relevant paths, whereas a large hop limit can introduce weakly related or noisy connections. The results demonstrate that predictive performance improves from $h=1$ to $h=3$, with Macro F1 rising from 0.7960 to 0.8197 and sensitivity increasing from 0.6732 to 0.7128. Extending the search to $h=4$ hops results in a slight degradation in sensitivity, confirming that over-expansion introduces distracting or irrelevant evidence to the LLM. Consequently, setting $h=3$ strikes the most effective practical balance between evidence coverage and noise mitigation in trajectory retrieval.


\begin{table}[t]
\centering
\caption{Hop-limit analyses on MIMIC-III mortality prediction.}
\label{tab:design_compact}
\resizebox{\columnwidth}{!}{
\begin{tabular}{llcccc}
\toprule
Analysis & Method & Macro-F1 & Sens. & Spec. & Acc. \\
\midrule
\multirow{4}{*}{Hop limit}
& 1-hop & 0.7960 & 0.6732 & 0.9471 & 0.9177 \\
& 2-hop & 0.8083 & 0.6930 & 0.9506 & 0.9229 \\
& 3-hop & \textbf{0.8197} & \textbf{0.7128} & 0.9531 & 0.9275 \\
& 4-hop & 0.8194 & 0.7029 & \textbf{0.9542} & \textbf{0.9291} \\
\bottomrule
\end{tabular}}
\end{table}

\section{Conclusion}

In this paper, we propose \textbf{TRACER}, a severity-aware trajectory RAG framework designed to enhance mortality and readmission prediction. By harnessing the reasoning capabilities of LLMs, TRACER enriches a patient's visit sequence with highly relevant medical context. Central to this framework is our severity-weighted medical KG, which assigns literature-backed severity scores to diagnosis concepts, thereby emphasizing high-risk clinical signals. We demonstrate the effectiveness of augmenting a patient's history with three critical dimensions of the patient: (1) severity-aware clinical trajectories that can capture disease progression across visits, (2) similar peer cases based on recent visit patterns, and (3) contextual evidence from unstructured clinical notes. This integration significantly improves both predictive performance and clinical transparency. Extensive experiments on MIMIC-III and MIMIC-IV demonstrate that TRACER consistently outperforms state-of-the-art baselines in sensitivity, Macro F1, specificity, and accuracy, offering a robust solution for interpretable, data-driven clinical risk assessment.

\begin{acks}
This work was partly supported by Institute of Information \& communications Technology Planning \& Evaluation (IITP) grant funded by the Korea government(MSIT) (No.RS-2020-II201373, Artificial Intelligence Graduate School Program(Hanyang University)) and the National Research Foundation of Korea(NRF) grant funded
by the Korea government(MSIT) (RS-2026-25487321), to K. Jeon, Y. Ko, and H. Kim. The work of W. Jung was supported by the National R\&D Program for Cancer Control through the National Cancer Center(NCC) funded by the Ministry of Health \& Welfare, Republic of Korea (RS-2025-02264000).
\end{acks}

\newpage
\bibliographystyle{ACM-Reference-Format}
\bibliography{references}

\newpage
\appendix
\section{Implementation Details}
\label{dataset_det}

We fix the random seed to 42 for all data splits, sampling steps, and retrieval operations to ensure reproducibility. For LLM prompting, we use OpenAI GPT-4o-mini with a temperature of 0.7 and a top-$p$ value of 1.0 in a zero-shot configuration.


All text prompts are constructed using a custom template engine, and are passed as raw strings to the API. The total length of each prompt is capped at 2,048 tokens; we prioritize risk trajectories and truncate protective ones if the patient's key supporting trajectories and her/his relevant clinical note passages exceed this limit. All embeddings computed by BioClinicalBERT and SimCSE are precomputed and cached to accelerate repeated inference.


During inference, we process one patient at a time due to LLM API latency constraints. A prediction for each patient takes approximately 4–6 seconds on average. All hyperparameters were tuned manually on the validation set, prioritizing sensitivity over Macro F1 score in alignment with clinical risk assessment guidelines. 


\subsection{Errors in Implementation of Label Definition:} 

We discovered critical errors in the preprocessing modules implemented in the PyHealth framework, which were used by several baselines such as KARE and GraphCare. These errors affected the construction of supervision signals for both the mortality and readmission prediction tasks.

For the mortality prediction task, patient's visits are initially sorted using \texttt{ROW\_ID} instead of the correct timestamp field \texttt{DISCHTIME} (discharge time). This misordering causes visit sequences to appear in a non-chronological order, leading to incorrect alignment between hospital visits and subsequent mortality events. As a result, some mortality labels are mistakenly attached to earlier visits rather than the final one preceding death.

For the readmission prediction task, prior models mistakenly use only \texttt{ADMITTIME} (admission time) when determining the time gap between successive visits, without properly considering the \texttt{DISCHTIME} (discharge time) of the preceding visit. This leads to misclassification of whether a readmission occurred within the clinically meaningful 15-day window, particularly in cases where two hospitalizations occurred in close succession.

To resolve these issues, we revise the preprocessing pipeline to: (1) strictly sorting visits by \texttt{DISCHTIME} for temporal consistency, and (2) computing readmission intervals based on the time difference between \texttt{DISCHTIME} (discharge time) of the current visit and \texttt{ADMITTIME} (admission time) of the next visit.

All mortality and readmission labels are recomputed accordingly. For example, in the MIMIC-III dataset, the proportion of positive labels changed from 5.42\% to 10.83\% for mortality prediction and from 54.82\% to 21.60\% for readmission prediction. Similar adjustments were observed in MIMIC-IV (mortality: 19.16\% $\rightarrow$ 38.13\%, readmission: 46.50\% $\rightarrow$ 48.84\%). All reported baseline results are based on the corrected label-definition protocol described above. After correcting the PyHealth label-construction logic, we regenerated the cohorts, reran all baselines, and retuned their hyperparameters under the corrected preprocessing. Thus, TRACER and all baselines use the same patient splits, labels, and preprocessing pipeline for fair comparison.

\section{Baselines}\label{appendix_baselines}
We compare \textbf{TRACER} with a range of representative baselines, spanning from early temporal models to recent models based on knowledge-enhanced LLMs. These baselines can be grouped into three main categories.

\subsection{Temporal Sequence Models}

\textbf{GRU} (2014): A gated recurrent neural network used for modeling visit sequences. It captures short-term temporal dependencies but lacks interpretability and long-range memory.

\noindent \textbf{Transformer} (2017): A self-attention-based model enabling long-range dependency modeling. While powerful, it does not explicitly encode medical conceptual hierarchies or relationships among medical concepts, nor does it account for their severity.

\noindent \textbf{RETAIN} (2016): An interpretable reverse-time attention model that highlights past medical events contributing to predictions. However, it does not incorporate external knowledge or concept severity.

\noindent \textbf{Deepr} (2016): Applies convolutional architectures to embed medical records but struggles to model long-term temporal progression.

\noindent \textbf{TCN} (2018): Temporal convolutional network capturing sequential information with causal dilated convolutions, though it ignores domain-specific relationships between medical concepts.

\subsection{Knowledge Graph-based Models}

\textbf{GRAM} (2017): This model introduces
hierarchical medical concept representations using an ontology-derived knowledge graph and attention mechanisms.

\noindent \textbf{ConCare} (2020): It models patient health context by capturing clinical patterns via attention across multiple temporal resolutions.

\noindent \textbf{AdaCare} (2020): It employs adaptive temporal attention to model irregular visit intervals in EHR sequences.

\noindent \textbf{GRASP} (2021): This model learns health representations by constructing a graph of similar patients and applies graph neural networks to transfer information across patients, enabling the model to learn clinical patterns across the population.

\noindent \textbf{StageNet} (2020): It models disease progression as latent stages, offering stage-aware attention mechanisms for prediction.

\noindent \textbf{GraphCare} (2024): It constructs patient-specific KGs using diagnoses, procedures, and medications. However, it relies on 1-hop expansion, only including concepts directly connected to the patient's medical codes and does not consider the severity of medical concepts.

\noindent \textbf{RAM-EHR} (2024): This model applies
retrieval-augmented generation to structured EHR data, enhancing prediction through relevant case-based memory but lacking fine-grained KG traversal.

\subsection{Text and LLM-based Models}

\noindent \textbf{MedRetriever} (2021): A retrieval-based framework leveraging unstructured medical text (e.g., clinical literature) for interpretable health risk prediction.

\begin{table*}[t]
\centering
\caption{Dataset Statistics of MIMIC-III and MIMIC-IV}
\label{tab:dataset_stats}
\resizebox{\textwidth}{!}{%
\begin{tabular}{lcccccccccccc}
\toprule
\textbf{Category} & \multicolumn{3}{c}{\textbf{MIMIC-III Mortality}} & \multicolumn{3}{c}{\textbf{MIMIC-III Readmission}} & \multicolumn{3}{c}{\textbf{MIMIC-IV Mortality}} & \multicolumn{3}{c}{\textbf{MIMIC-IV Readmission}} \\
\cmidrule(lr){2-4} \cmidrule(lr){5-7} \cmidrule(lr){8-10} \cmidrule(lr){11-13}
& Train & Valid & Test & Train & Valid & Test & Train & Valid & Test & Train & Valid & Test \\
\midrule
\# Patients & 7686 & 964 & 932 & 7608 & 997 & 972 & 7962 & 1002 & 1036 & 7976 & 990 & 1034 \\
\# Visits / Patient & 1.86 & 2.11 & 1.77 & 1.84 & 1.93 & 2.11 & 1.60 & 1.66 & 1.63 & 1.37 & 1.36 & 1.32 \\
\# Diagnosis / Patient & 21.68 & 23.04 & 21.82 & 21.56 & 22.86 & 22.89 & 19.86 & 21.03 & 20.44 & 14.89 & 15.12 & 14.87 \\
\# Procedures / Patient & 6.76 & 6.61 & 6.96 & 6.72 & 6.84 & 6.96 & 4.06 & 4.16 & 4.12 & 3.40 & 3.31 & 3.37 \\
\# Medications / Patient & 53.15 & 60.37 & 53.61 & 53.10 & 56.31 & 57.97 & 42.82 & 44.71 & 43.68 & 33.13 & 33.01 & 32.94 \\
\bottomrule
\end{tabular}%
}
\end{table*}

\noindent \textbf{KARE} (2025): A retrieval-augmented generation model that constructs reasoning chains over retrieved knowledge and textual evidence using LLMs. However, it uses iterative reasoning chain generation, which incurs high latency and lacks explicit severity modeling.

\noindent \textbf{DearLLM} (2025): This model enhances
patient representations by querying LLMs to infer correlation strength between medical concepts, but lacks structured graph traversal and multi-modal context aggregation.

\section{Dataset}\label{dataset_stats}

We utilize two widely adopted and publicly available electronic health record (EHR) datasets: \textbf{MIMIC-III v1.4} and \textbf{MIMIC-IV v2.0}. Both datasets are hosted on PhysioNet and accessible under strict data use agreements that enforce de-identification and responsible research conduct. Dataset statistics are summarized in Table~\ref{tab:dataset_stats}.

\subsection{Patient Privacy and Ethical Considerations}

All data used in this study are fully de-identified in compliance with the Health Insurance Portability and Accountability Act (HIPAA) Safe Harbor standards. No personally identifiable information is retained or accessible. Additionally:
\begin{itemize}
    \item Only researchers certified via CITI training and approved by PhysioNet may access the datasets.
    \item We do not introduce any additional linkage or re-identification processes.
    \item Our code, models, and processed results do not expose individual-level information, and all reporting is conducted at an aggregate level.
\end{itemize}

We further adopt \emph{split-by-patient} strategies to ensure that no leakage occurs between training and test sets, and that each patient appears in only one split.

\subsection{Dataset Preprocessing}

We follow established preprocessing pipelines aligned with prior works such as GraphCare and PyHealth, ensuring reproducibility and compatibility.

\noindent \textbf{Inclusion criteria}: In the reported experiments, we retain only patients with at least two hospital visits. For each retained patient, the last visit is used as the prediction target, and all preceding visits are used as historical context for training and inference. Accordingly, a patient with exactly two visits contributes one historical visit and one target visit, while single-visit patients are excluded from the main experimental cohort. We report the resulting cohort sizes after this filtering step.

\noindent \textbf{Medical concept mapping}: We mapped medical concepts (i.e., diagnosis, procedure, medication) by using the following classification systems.
\begin{itemize}
    \item Diagnoses and procedures are mapped using Clinical Classifications Software (CCS).
    \item Medications are mapped to the third level of the Anatomical Therapeutic Chemical classification (ATC-3)\footnote{https://www.who.int/tools/atc-ddd-toolkit/atc-classification}.
\end{itemize}

\noindent \textbf{Visit truncation}: We exclude patients with more than 10 visits to prevent outlier effects and manage computational complexity.

\subsection{Sampling Details}

\noindent \textbf{MIMIC-III}: The full dataset is used with the same split strategy as KARE. Mortality and readmission labels are derived from visit-level transitions.

\noindent \textbf{MIMIC-IV Mortality}: We retain 2,152 positive samples (label = 1) and sample 10,000 negative patients (label = 0) who have $\leq$10 visits, ensuring class imbalance mitigation.

\noindent \textbf{MIMIC-IV Readmission}: We randomly select 5,000 readmitted patients and 5,000 non-readmitted patients.

All subsets are split into training/validation/test sets in an 8:1:1 ratio, ensuring:
\begin{itemize}
    \item No overlap of patients between splits (split-by-patient).
    \item Balanced representation of outcomes across subsets.
\end{itemize}

\section{Evaluation Metrics}\label{appendix_metrics}
We evaluate model performance using four key metrics: \textbf{Macro F1}, \textbf{Sensitivity (Recall)}, \textbf{Specificity}, and \textbf{Accuracy}. Each metric reflects a different aspect of clinical utility and decision-making reliability, particularly in high-stakes settings such as mortality or readmission prediction.

\noindent\textbf{Sensitivity (Recall)}: Sensitivity quantifies the proportion of actual positive cases 
(i.e., patients who will die in the mortality prediction task; those who will be readmitted in the readmission task) that are correctly identified by the model. In clinical applications, high sensitivity is crucial to minimize false negatives, i.e., cases where the model incorrectly predicts a patient as low-risk when they are actually high-risk. Such failures can have serious consequences, including missed ICU triage, delayed interventions, or preventable deterioration. For example, predicting a high-risk mortality case as ``survival'' could result in the patient not receiving necessary monitoring or escalation of care. Therefore, we prioritize sensitivity in evaluation, especially given the high class imbalance typical in healthcare datasets.

\begin{equation}
\text{Sensitivity} = \frac{\text{TP}}{\text{TP} + \text{FN}}
\end{equation}

\begin{table*}[t]
\centering
\caption{Overall performance of mortality and readmission prediction on the MIMIC-III dataset. The numbers in parentheses indicate the proportion of patients with positive labels (label=1) for each task. The best and second best results are denoted in bold and underlined, respectively.}
\label{tab:main_results_mimic3}
\begin{tabular}{lcccccccc}
\toprule
\textbf{Models} & \multicolumn{4}{c}{\textbf{Mortality Prediction (10.83\%)}} & \multicolumn{4}{c}{\textbf{Readmission Prediction (21.6\%)}} \\
\cmidrule(lr){2-5} \cmidrule(lr){6-9}
& Accuracy & Macro F1 & Sensitivity & Specificity & Accuracy & Macro F1 & Sensitivity & Specificity \\
\midrule
GRU (2014) & 0.8605 & 0.5349 & 0.1089 & 0.9520 & 0.7368 & 0.5075 & 0.1238 & 0.8901 \\
Transformer (2017) & 0.8637 & 0.5369 & 0.1089 & 0.9558 & 0.7335 & 0.5146 & 0.1428 & 0.8987 \\
RETAIN (2016) & 0.8583 & 0.5163 & 0.0594 & 0.9504 & 0.7495 & 0.5653 & 0.1744 & 0.9096 \\
GRAM (2017) & 0.8594 & 0.5309 & 0.1188 & 0.9494 & 0.7510 & 0.5266 & 0.1438 & 0.9125 \\
Deepr (2016) & 0.8487 & 0.5262 & 0.1004 & 0.9522 & 0.7386 & 0.5342 & 0.1494 & 0.9175 \\
TCN (2018) & 0.7588 & 0.4904 & 0.1515 & 0.8329 & 0.7731 & 0.5317 & 0.1264 & 0.9501 \\
ConCare (2020) & 0.8920 & 0.4715 & 0.0734 & 0.9602 & 0.7308 & 0.4983 & 0.1267 & 0.8904 \\
AdaCare (2020) & 0.8328 & 0.5130 & 0.0990 & 0.9621 & 0.7315 & 0.5025 & 0.1571 & 0.8898 \\
GRASP (2021) & 0.8938 & 0.5176 & 0.0845 & \textbf{0.9963} & 0.7335 & 0.5260 & 0.1268 & 0.9056 \\
StageNet (2020) & 0.8390 & 0.5571 & 0.1881 & 0.9189 & 0.7395 & 0.5275 & 0.1962 & 0.8923 \\
GraphCare (2024) & 0.9229 & 0.5963 & 0.2267 & 0.9577 & 0.7599 & 0.6364 & 0.6999 & 0.9567 \\
RAM-EHR (2024) & 0.8852 & 0.6105 & 0.2277 & 0.9627 & 0.7451 & 0.5772 & 0.2574 & 0.9562 \\
KARE (2025) & \underline{0.9227} & \underline{0.6375} & 0.3711 & 0.9593 & 0.7724 & 0.7271 & 0.7531 & 0.7162 \\
MedRetriever (2021) & 0.8358 & 0.5622 & 0.2079 & 0.9612 & 0.7562 & 0.5970 & 0.6524 & 0.9168 \\
DearLLM (2025) & 0.8683 & 0.6374 & 0.2673 & 0.9615 & 0.7343 & 0.6390 & 0.7095 & 0.9163 \\
\textbf{Ours} & \textbf{0.9334} & \textbf{0.8284} & \textbf{0.6831} & \textbf{0.9639} & \textbf{0.9135} & \textbf{0.8709} & \textbf{0.7809} & \textbf{0.9501} \\
\bottomrule
\end{tabular}
\end{table*}

\begin{table*}[t]
\centering
\caption{Overall performance of mortality and readmission prediction on the \miv dataset. The numbers in parentheses indicate the proportion of patients with positive labels (label=1) for each task. The best and second best results are denoted in bold and underlined, respectively.}
\label{tab:main_results_mimic4}
\begin{tabular}{lcccccccc}
\toprule
\textbf{Models} & \multicolumn{4}{c}{\textbf{Mortality Prediction (38.13\%)}} & \multicolumn{4}{c}{\textbf{Readmission Prediction (48.84\%)}} \\
\cmidrule(lr){2-5} \cmidrule(lr){6-9}
& Accuracy & Macro F1 & Sensitivity & Specificity & Accuracy & Macro F1 & Sensitivity & Specificity \\
\midrule
GRU (2014) & 0.7722 & 0.7220 & 0.4557 & 0.9672 & 0.6209 & 0.5558 & 0.6891 & 0.5558 \\
Transformer (2017) & 0.7780 & 0.7377 & 0.5063 & 0.9454 & 0.6122 & 0.6162 & 0.6495 & 0.5766 \\
RETAIN (2016) & 0.7722 & 0.7389 & 0.5142 & 0.9126 & 0.6196 & 0.6256 & 0.6564 & 0.6957 \\
GRAM (2017) & 0.7751 & 0.7400 & 0.5342 & 0.9236 & 0.6284 & 0.6271 & 0.6911 & 0.5671 \\
Deepr (2016) & 0.7983 & 0.7627 & 0.5392 & 0.9159 & 0.6313 & 0.6315 & 0.6897 & 0.5916 \\
TCN (2018) & 0.7963 & 0.7560 & 0.5146 & 0.9719 & 0.6181 & 0.6175 & 0.6156 & 0.6256 \\
ConCare (2020) & 0.7971 & 0.7573 & 0.5114 & 0.9758 & 0.6074 & 0.6072 & 0.6436 & 0.5728 \\
AdaCare (2020) & 0.8058 & 0.7717 & 0.5468 & 0.9657 & 0.6278 & 0.6337 & 0.7032 & 0.5924 \\
GRASP (2021) & 0.7925 & 0.7514 & 0.5063 & 0.9688 & 0.5881 & 0.5876 & 0.5980 & 0.5766 \\
StageNet (2020) & 0.7809 & 0.7479 & 0.5146 & 0.9560 & 0.6096 & 0.5904 & 0.5894 & 0.5622 \\
GraphCare (2024) & 0.8782 & 0.8356 & 0.6327 & 0.9491 & 0.6493 & 0.6713 & 0.6831 & 0.6763 \\
RAM-EHR (2024) & 0.8460 & 0.7492 & 0.6179 & 0.9708 & 0.6173 & 0.6543 & 0.7052 & 0.6564 \\
KARE (2025) & 0.8893 & 0.8698 & 0.7498 & 0.9771 & 0.7236 & 0.7444 & 0.8582 & 0.6469 \\
MedRetriever (2021) & 0.8021 & 0.7727 & 0.5797 & 0.9392 & 0.6247 & 0.6237 & 0.6901 & 0.6525 \\
DearLLM (2025) & 0.8349 & 0.8143 & 0.6456 & 0.9589 & 0.6876 & 0.6987 & 0.7485 & 0.6295 \\
\textbf{Ours} & \textbf{0.8938} & \textbf{0.8830} & \textbf{0.7722} & \textbf{0.9688} & \textbf{0.7949} & \textbf{0.7942} & \textbf{0.8752} & \textbf{0.7183} \\
\bottomrule
\end{tabular}
\end{table*}

\textbf{Macro F1 Score}: Macro F1 is the unweighted average of F1 scores computed separately for the positive and negative classes. This metric is particularly suited to imbalanced classification problems, where the majority class (i.e., survivors in the mortality prediction task; non-readmitted patients in the readmission task) 
dominates. By averaging performance across classes, Macro F1 ensures that the model’s performance on minority classes (i.e., death or readmission) is not overshadowed. In clinical risk prediction, this reflects the ethical need to treat false reassurance and over-alarm symmetrically, and to maintain balance between over- and under-treatment.

\begin{equation}
\text{F1}_{\text{class}} = \frac{2 \cdot \text{Precision} \cdot \text{Recall}}{\text{Precision} + \text{Recall}}, \quad
\text{Macro F1} = \frac{1}{2}(\text{F1}_{\text{positive}} + \text{F1}_{\text{negative}})
\end{equation}

\noindent\textbf{Specificity}: Specificity measures the proportion of true negative cases 
(i.e., patients who survive in the mortality prediction task; those who are not readmitted in the readmission task) that are correctly identified. While high sensitivity is critical to avoid missed risk, high specificity is also desirable to avoid false positives, which can lead to unnecessary interventions, patient anxiety, or resource waste. For instance, a false alarm on mortality may result in aggressive care or ICU admission for patients who are actually stable. Therefore, we report specificity alongside sensitivity to evaluate the trade-off between caution and overreaction.

\begin{equation}
\text{Specificity} = \frac{\text{TN}}{\text{TN} + \text{FP}}
\end{equation}

\noindent\textbf{Accuracy}: Accuracy measures the overall proportion of correct predictions, combining both positive and negative cases. However, in highly imbalanced datasets such as MIMIC-III and MIMIC-IV (e.g., 10.83\% mortality rate), accuracy can be misleading. A naive classifier that predicts all patients as negative could still achieve high accuracy while completely failing to detect true positives. Therefore, we report accuracy only as a reference metric, not as the primary indicator of clinical utility.

\begin{equation}
\text{Accuracy} = \frac{\text{TP} + \text{TN}}{\text{TP} + \text{TN} + \text{FP} + \text{FN}}
\end{equation}

In real-world hospital settings, the cost of misclassifying a critical patient (false negative) is generally much higher than that of a false alert (false positive). For example, ICU beds are limited resources, and failing to identify a high-risk case could result in missed opportunities for life-saving care. Thus, a model with high sensitivity and balanced Macro F1 is more aligned with the operational priorities of clinical decision support systems (CDSS). Our evaluation framework reflects this by placing particular emphasis on these two metrics.

\section{Main Results}\label{append:main_results}
We report the main results of our experiments on both the MIMIC-III and MIMIC-IV datasets for mortality and 15-day readmission prediction tasks. \textbf{TRACER} consistently outperforms all baselines across all evaluation metrics, including Macro F1, sensitivity, specificity, and accuracy.

Table \ref{tab:main_results_mimic3} and Table \ref{tab:main_results_mimic4} present the quantitative comparisons on MIMIC-III and MIMIC-IV, respectively. The best and second-best results are denoted in bold and underlined, respectively.

\subsection{Mortality Prediction}

\textbf{TRACER} achieves substantial gains in sensitivity---up to 84.1\% improvement on MIMIC-III and 29.5\% improvement on MIMIC-IV compared to the strongest baseline, KARE. These improvements are clinically meaningful, as higher sensitivity corresponds to fewer missed high-risk patients, reducing the chance of avoidable deaths.

\textbf{TRACER} also leads in Macro F1 score, indicating that it balances performance across the minority (death) and majority (survival) classes better than all competitors. This balance is critical in clinical settings to prevent both over- and under-treatment.

We attribute these gains to the following core innovations in TRACER:
\begin{itemize}
    \item \textbf{Severity-aware trajectory retrieval} from the medical knowledge graph allows for richer representations of disease progression and patient deterioration than flat paths or 1-hop neighborhoods used in GRASP or GraphCare.
    \item \textbf{Repacking and refinement techniques} enhance LLM attention on the most informative paths and notes, mitigating the ``lost-in-the-middle'' effect common in long LLM prompts.
    \item \textbf{Patient similarity-based augmentation} enables LLMs to reason by analogy to prior cases with similar disease trajectories, increasing prediction robustness.
\end{itemize}

\subsection{Readmission Prediction}

While the performance gap is less dramatic than mortality, \textbf{TRACER} still shows clear superiority on all metrics. For instance, on MIMIC-IV, TRACER improves specificity by $\geq$15.5\% over the strongest baseline. This is crucial in practice to avoid excessive hospital readmissions, which are costly and often unnecessary.

Performance gains stem largely from:
\begin{itemize}
    \item \textbf{Salient clinical note retrieval}, which captures discharge planning quality, unresolved symptoms, and social context, i.e., factors often missing in structured EHR data.
    \item \textbf{Integration of protective trajectories}, which help the model distinguish between temporary and persistent risk factors by tracing longitudinal improvements in patient status and identifying interventions that have successfully reversed clinical deterioration.
\end{itemize}

\paragraph{Summary}

In both tasks, the performance of \textbf{TRACER} demonstrates the value of its hybrid retrieval-augmented design. The integration of structured knowledge (SMKG), unstructured notes, and similar patient context, paired with severity-aware trajectory modeling, enables not only accurate but also interpretable and clinically relevant predictions.

These results validate our hypothesis that capturing causal progression and amplifying high-risk signals, rather than flattening medical history, leads to more effective risk stratification in real-world clinical data.

\section{Robustness to Data Sparsity across Visit Sequence Lengths}\label{append:intro_2}



\begin{figure}[t]
    \centering
    \includegraphics[width=1\columnwidth]{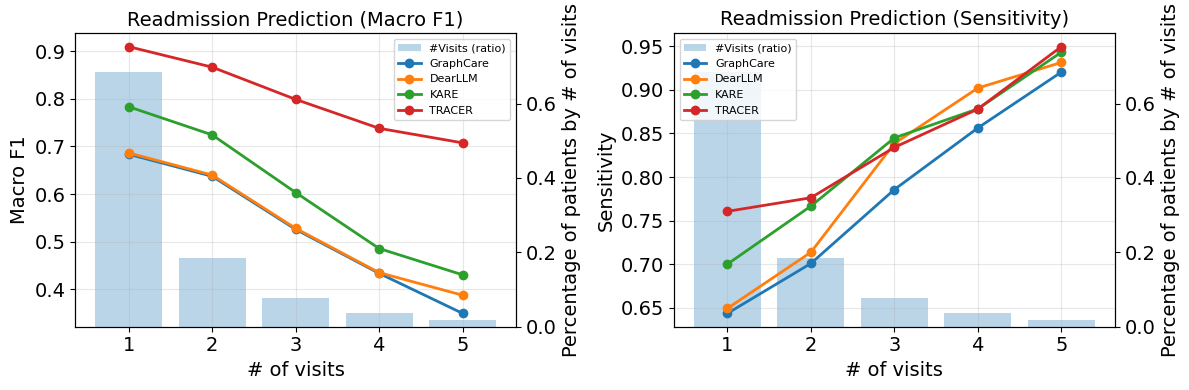}
    \caption{Performance in terms of Macro F1 and sensitivity across visit sequence lengths for the readmission prediction task, with the percentage of patients by number of visits.
    }
    \label{fig:intro2}
\end{figure}

We analyze the robustness under increasing data sparsity that arises as visit sequences become longer. Patients are divided into five groups according to the exact number of visits: 1, 2, 3, 4, and 5. Figure \ref{fig:intro2} illustrates the performance trends across visit lengths, where the number of visits is used as the horizontal axis, the line plots present the Macro F1 and sensitivity of the baselines and TRACER, and the background bars represent the percentage of patients associated with each visit length.

Most baselines exhibit a steep decline in Macro F1 as the number of visits increases, indicating that their performance degrades rapidly for patients with longer visit sequences. In contrast, TRACER shows a much more gradual decrease in Macro F1, maintaining comparatively stable performance even in the longer-visit groups. A similar trend is observed for sensitivity: as the number of visits increases, TRACER demonstrates a smooth and consistent increase, whereas baselines exhibit more abrupt or unstable changes.

These results show that TRACER is more robust to the data sparsity that arises with longer visit sequences. By leveraging trajectories of similar patients selected based on the target patient’s recent visits, TRACER focuses on the most relevant and up-to-date clinical status rather than simply accumulating historical records. This design enables more stable and reliable predictions for patients with longer visit histories.

\section{Impact of Repacking and Refinment}\label{ablation2}
Table \ref{tab:abl2} validates the effectiveness of trajectory refinement and repacking. In the variant "\textit{w/o} Traj. Ref.", all shortest paths with up to three hops between concepts in adjacent visits form the trajectories. In the “\textit{w/o} Traj. Rep.” variant, all trajectories are randomly shuffled prior to input to the LLM. In contrast, trajectory refinement selects only the most important paths based on the MMR score. Experimental results show that the removal of trajectory refinement decreases the sensitivity of mortality prediction by 7.2\% (0.6831 $\rightarrow$ 0.6336).

\begin{table}[t]
\centering
\caption{Ablation study on reranking and repacking in TRACER}
\label{tab:abl2}
{\fontsize{8}{8}\selectfont
\begin{tabular}{lcccc}
\toprule
\multirow{2}{*}{\textbf{Models}} 
  & \multicolumn{2}{c}{\textbf{Mortality (10.83\%)}} 
  & \multicolumn{2}{c}{\textbf{Readmission (21.6\%)}} \\
\cmidrule(lr){2-3} \cmidrule(lr){4-5}
 & Macro F1 & Sensitivity & Macro F1 & Sensitivity \\
\midrule
TRACER                         & \textbf{0.8284} & \textbf{0.6831} & \textbf{0.8709} & \textbf{0.7809} \\
\textit{w/o} Traj. Ref.      & 0.7944 & 0.6336 & 0.8469 & 0.7476 \\
\textit{w/o} Traj. Rep.      & 0.8097 & 0.6435 & 0.8626 & 0.7714 \\
\textit{w/o} Pat. Rep.  & 0.8153 & 0.6534 & 0.8604 & 0.7571 \\
\bottomrule
\end{tabular}%
} 
\end{table}

This demonstrates that MMR-based refinement has a significant positive effect on performance. Furthermore, omitting trajectory repacking results in a 5.8\% decrease in sensitivity (0.6831 $\rightarrow$ 0.6435), while the removal of similar patient repacking leads to a 4.3\% decrease (0.6831 $\rightarrow$ 0.6534). This demonstrates that repacking trajectories is more effective than repacking similar patients, indicating that the LLM is more sensitive to the order and arrangement of target patient’s trajectories. Thus, careful selection and ordering of trajectories are essential for optimal model performance.

\begin{table}[t]
\centering
\caption{Effect of severity score granularity on mortality prediction
performance on the MIMIC-III dataset.}
\label{tab:abl3}
{\fontsize{7}{7}\selectfont
\begin{tabular}{lcccc}
\toprule
\textbf{Severity Scale} & \textbf{Macro F1} & \textbf{Sensitivity} & \textbf{Specificity} & \textbf{Accuracy} \\
\midrule
1--100 (5-level prompt) & \textbf{0.8284} & 0.6831 & \textbf{0.9639} & \textbf{0.9334} \\
1--20                  & 0.8197 & \textbf{0.7128} & 0.9531 & 0.9275 \\
1--5                   & 0.7271 & 0.5842 & 0.9194 & 0.8831 \\
\bottomrule
\end{tabular}
}
\vspace{-10pt}
\end{table}

\begin{figure*}[t]
    \centering
    \includegraphics[width=0.8\textwidth]{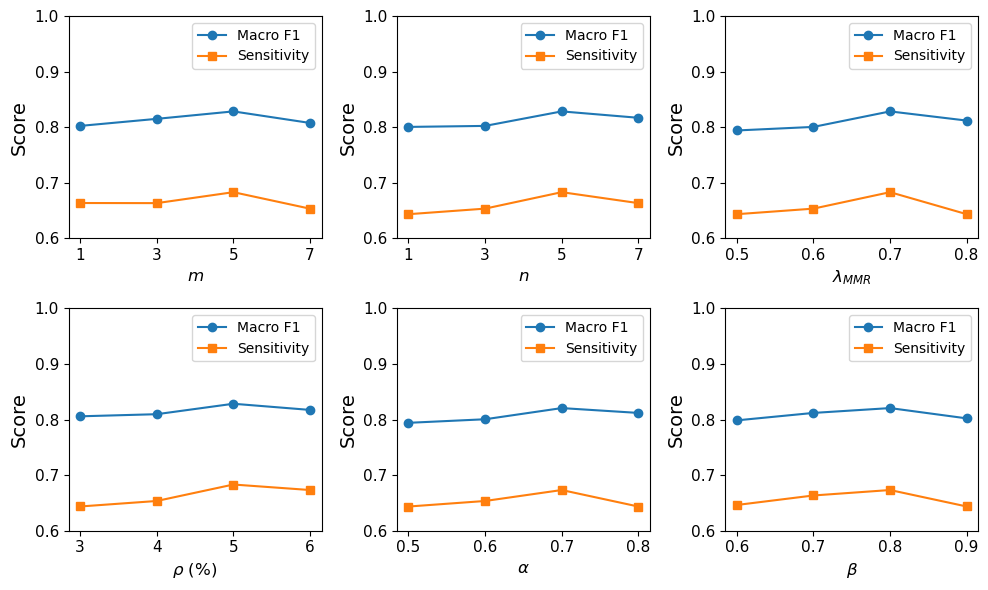}
    \vspace{-10pt}
    \caption{Hyperparameter sensitivity analysis of the number of similar patients, the number of clinical notes, the proportion of trajectories, and the weight for the MMR score on MIMIC-III.}
    \label{fig:hyperparameters2}
\end{figure*}

\section{Impact of Severity Score Granularity.}\label{appendix_scoring}
We further investigate the impact of severity score granularity on mortality prediction performance. Specifically, we compare three scoring schemes: a fine-grained severity score in the range of 1--100 (implemented as a 5-level prompt), a moderately coarse
scheme in the range of 1--20, and a coarse-grained scheme in the
range of 1--5.

As shown in Table \ref{tab:abl3}, the 1 --100 severity score achieves the best performance in terms of Macro F1, specificity, and accuracy, while the 1--20 scheme yields the highest sensitivity. In contrast, the 1--5 scheme consistently underperforms
across all metrics, with a particularly notable drop in Macro F1 and sensitivity.
This performance degradation suggests that coarse-grained scoring fails to capture meaningful differences among patients with varying levels of clinical severity. In additional experiments, we
observe that severity scores are more densely distributed in the 51--100 range than in the 1--50 range, indicating that fine-grained scores better differentiate high-risk patients. This observation
aligns with the design motivation of clinical risk stratification and is consistent with prior severity indices, such as the Charlson Comorbidity Index (CCI), which also emphasizes higher-weight severe conditions.
These results indicate that fine-grained severity modeling is crucial for accurately capturing disease severity and improving mortality prediction, and that overly coarse scoring schemes may obscure clinically important risk differences.

\section{Robustness Analysis}\label{appendix_robustness}

\begin{table}[t]
\centering
\caption{Comparison across cutting-edge clinical risk prediction baselines on MIMIC-IV. The best results are denoted in bold.}
\label{tab:comparison_clinical_risk_baselines_mimiciv}
{\fontsize{8}{8}\selectfont
\resizebox{\columnwidth}{!}{%
\begin{tabular}{lcccc}
\toprule
\textbf{Methods} & Macro F1 & Sensitivity & Specificity & Accuracy \\
\midrule
TRACER & \textbf{0.8830} & \textbf{0.7722} & \textbf{0.9688} & \textbf{0.8939} \\
ETHOS & 0.5892 & 0.7621 & 0.9002 & 0.8895 \\
Zero-shot LLM & 0.7996 & 0.6203 & 0.9501 & 0.8244 \\
\bottomrule
\end{tabular}%
}
}
\vspace{-10pt}
\end{table}

\begin{table}[t]
\centering
\caption{Robustness analysis across knowledge snapshots on MIMIC-III. The best results are denoted in bold.}
\label{tab:robustness_snapshots}
{\fontsize{8}{8}\selectfont
\resizebox{\columnwidth}{!}{%
\begin{tabular}{lcccc}
\toprule
\textbf{Methods} & Macro F1 & Sensitivity & Specificity & Accuracy \\
\midrule
Ours & \textbf{0.8197} & \textbf{0.7128} & \textbf{0.9531} & \textbf{0.9275} \\
PubMed $\sim$2024.12 & 0.8168 & 0.7029 & \textbf{0.9531} & 0.9267 \\
PubMed $\sim$2020.12 & 0.8066 & 0.6930 & 0.9494 & 0.9216 \\
API 2024-07-18 & 0.7843 & 0.6831 & 0.9374 & 0.9094 \\
\bottomrule
\end{tabular}%
}
}
\vspace{-10pt}
\end{table}

\subsection{Comparison across Cutting-edge Clinical Risk Prediction Baselines}\label{sec:appendix_main_result_baselines}

Table~\ref{tab:comparison_clinical_risk_baselines_mimiciv} compares TRACER with cutting-edge clinical risk prediction baselines on MIMIC-IV mortality prediction. TRACER outperforms ETHOS~\cite{ETHOS} and a prompt-only zero-shot LLM baseline~\cite{ZhuZeroShotEHR} across all reported metrics, including Macro F1, sensitivity, specificity, and accuracy. These results indicate that TRACER provides stronger predictive signals through retrieval-augmented evidence construction than direct clinical risk prediction baselines or prompt-only inference.

\subsection{Robustness across Knowledge Snapshots}\label{sec:appendix_robustness_snapshots}

Table~\ref{tab:robustness_snapshots} analyzes the robustness of TRACER under different knowledge and model snapshots on MIMIC-III mortality prediction. Compared with the full setting, older PubMed snapshots lead to only bounded degradation, and the API snapshot also preserves competitive performance. These results suggest that TRACER is robust to temporal variation in retrieved biomedical knowledge and model versions.

\section{Hyperparameter Sensitivity}\label{appendix_hyperparameters}

We conduct a comprehensive sensitivity analysis on six key hyperparameters in the \textbf{TRACER} framework, using the MIMIC-III dataset for the mortality prediction task (10.83\% positive class). We report accuracy, Macro F1, sensitivity, and specificity for each setting.

\subsection{Risk Trajectory Relevance Weight ($\alpha$)}

The parameter $\alpha$ controls the trade-off between semantic relevance and trajectory severity in risk trajectory retrieval. Performance peaks at $\alpha = 0.7$, yielding: Macro F1 score = 0.8284, sensitivity = 0.6831.

A smaller $\alpha$ underemphasizes severity, while a larger $\alpha$ overemphasizes it and reduces semantic alignment. Therefore, a balanced 
$\alpha=0.7$ provides optimal retrieval fidelity.

\subsection{Protective Trajectory Relevance Weight ($\beta$)}

This governs the same trade-off for protective trajectories. The best performance is observed at $\beta = 0.3$, with: Macro F1 score = 0.8282, sensitivity = 0.6831. This suggests that overemphasizing non-risky signals (e.g., $\beta = 0.4$) may dilute focus on high-risk contrastive reasoning.

\subsection{Number of Retrieved Trajectories ($k$)}

We vary the proportion of top-$k$ trajectories used per patient per factor (risk/protective). The best performance is achieved at $k = 5\%$, delivering: Macro F1 score = 0.8284, sensitivity = 0.6831. A lower $k$ (e.g., 3\%) reduces coverage, while a higher $k$ (e.g., 6\%) introduces redundancy and distracts the LLM.

\subsection{Number of Retrieved Clinical Notes ($n$)}

The number of retrieved clinical note passages significantly affects model robustness. $n = 5$ provides the best trade-off: Macro F1 score = 0.8284, sensitivity = 0.6831. Too few notes underrepresent the clinical context, and too many dilute key signals (e.g., $n = 7$ drops F1 to 0.8076).

\subsection{Number of Retrieved Similar Patients ($m$)}

Retrieving top-$m$ similar patients using recent-visit Jaccard similarity improves the case-based reasoning of LLM. $m = 5$ again yields optimal performance: Macro F1 score = 0.8284, sensitivity = 0.6831. This setting balances analogy and prompt length. Using too few (e.g., $m = 1$) or too many (e.g., $m = 7$) leads to underfitting or distraction, respectively.

\subsection{MMR Weight in Refinement ($\lambda_{\text{MMR}}$)}

This parameter governs the trade-off between relevance and diversity during trajectory refinement using Maximal Marginal Relevance (MMR). At $\lambda = 0.7$, we observe: Macro F1 score = 0.8284, sensitivity = 0.6831. When $\lambda$ is too low (e.g., 0.5), redundancy increases; when $\lambda$ is too high (e.g., 0.9), diversity is lost.

\section{Correlation Analysis between Severity Scores and Mortality Labels.}\label{appendix_correlation}

For in-depth analysis, we compute the point-biserial correlation coefficient between a patient-level severity score and mortality label for every patient. Since our severity scores are defined at the diagnosis level, we aggregate them into the patient-level severity score by averaging the severity scores of all diagnoses observed in the patient’s visit sequence, similar to the scoring technique of Charlson Comorbidity Index (CCI; normalized to a 37-point scale). The patient-level severity scores (normalized to a 100-point scale) achieve a correlation of 0.38 with the mortality labels, while the CCI shows a correlation of 0.29. Both correlations are statistically significant
with p-values below 0.05.
We observe that CCI tends to assign higher scores to patients with specific high-weight comorbid conditions, as it relies on predefined disease-specific weights. In contrast, the proposed severity score captures a broader range of clinical severity signals, which leads to a stronger overall association with mortality outcomes.
Notably, the mortality label in our dataset is highly imbalanced, with only 10.83\% positive cases. This label skew generally suppresses the magnitude of correlation coefficients, suggesting that the observed correlations may underestimate the true relationship
between clinical severity and mortality risk.

\clearpage
\begin{figure*}[t]
    \centering
    \includegraphics[width=\textwidth]{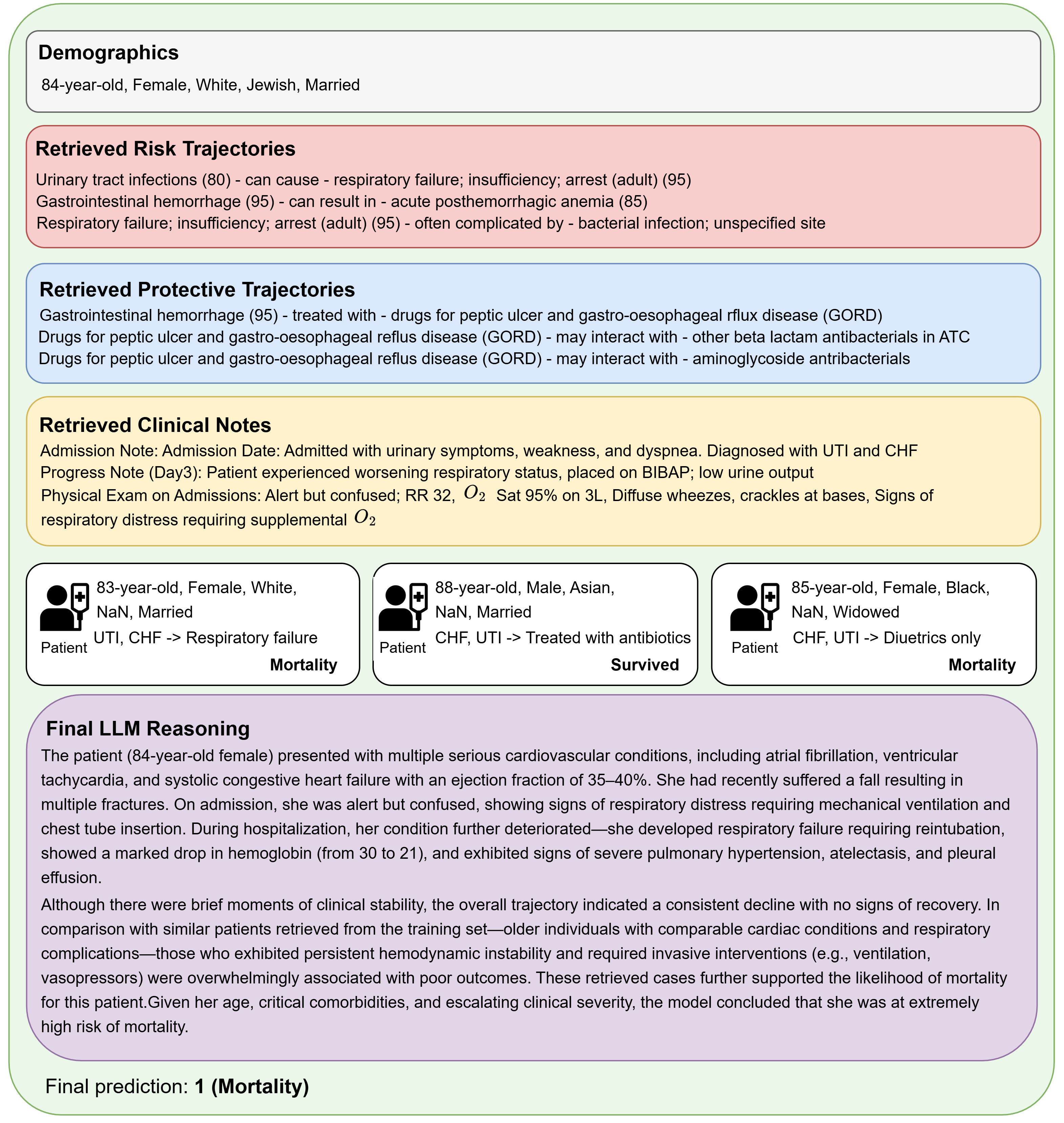}
    \caption{Case study on mortality prediction using TRACER.}
    \label{fig:tracer}
\end{figure*}
\clearpage

\begin{figure*}[t]
    \centering
    \includegraphics[width=\textwidth]{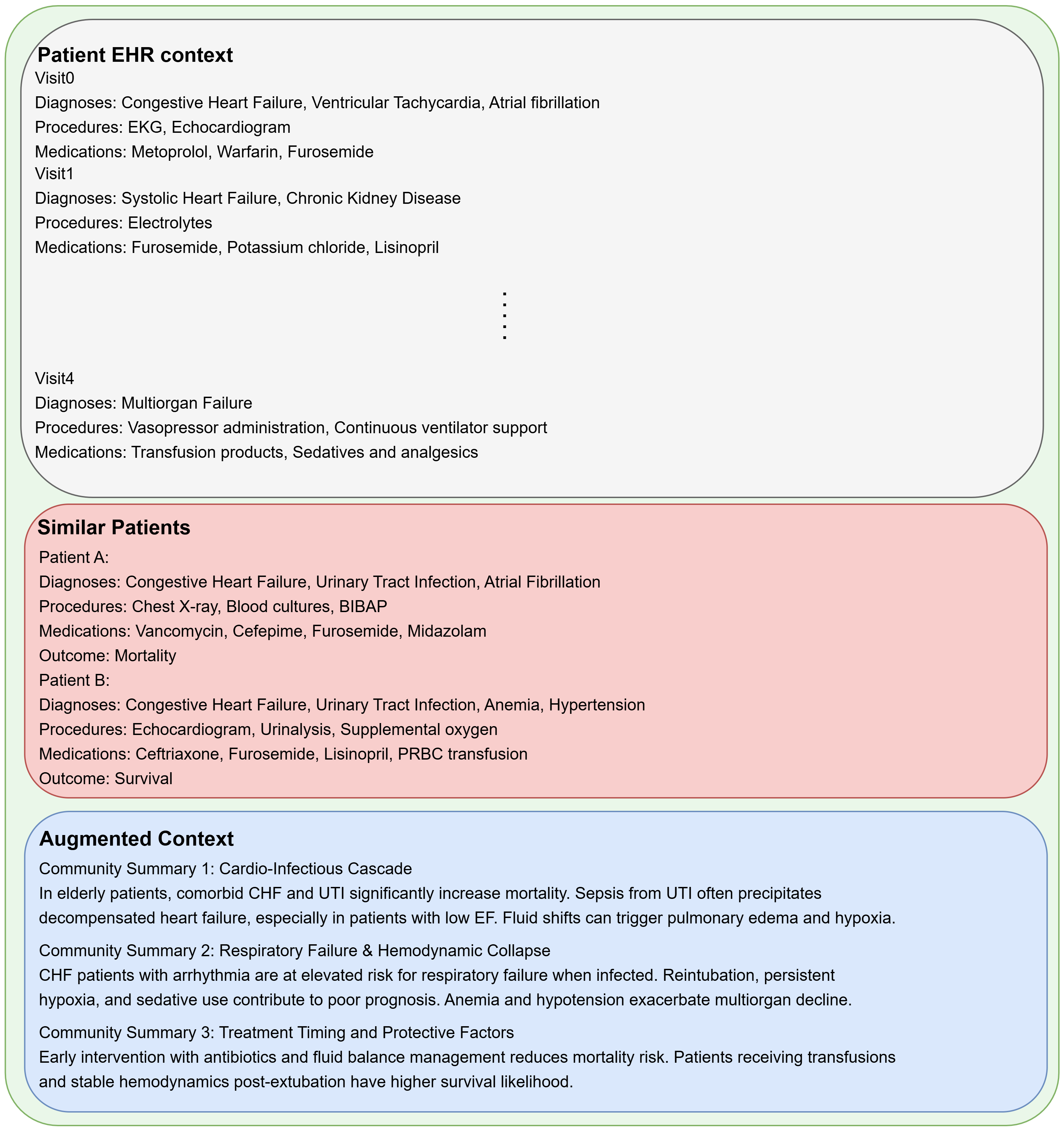}
    \caption{Case study on mortality prediction using other models.}
    \label{fig:kare}
\end{figure*}
\clearpage

\section{Case Study}
\subsection*{Comparison of Case Study Inputs Between TRACER and Other Models}
To highlight the representational differences between our proposed framework and prior approaches, we provide a side-by-side case study comparison using the same target patient. While both models aim to predict clinical risk based on EHR data, they differ in how contextual information is retrieved, represented, and integrated.

\textbf{Trajectory and Clinical Note Context}: Our framework explicitly retrieves risk and protective trajectories from a severity-weighted medical knowledge graph (SMKG), along with key evidence sentences from clinical notes. In the comparison figure (Figure \ref{fig:tracer}), we include representative key supporting trajectories (KSTs) and key sentences from the retrieved note to reflect what our model uses during inference. In contrast, other models (Figure \ref{fig:kare}) do not explicitly utilize trajectory-level knowledge or unstructured text. Instead, relevant clinical insights are implicitly captured through community-level knowledge graph summaries or structured EHR features. In Figure \ref{fig:tracer}, we annotate these regions accordingly to clarify how comparable information is handled.
    
\textbf{Similar Patient Representation}: In our framework, similar patients are selected based on recent visit patterns and demographic similarity, and are represented via concise trajectory-based summaries. These are presented as one-sentence descriptions in Figure \ref{fig:tracer}. Other models (Figure \ref{fig:kare}) incorporate exemplar patients as structured EHR instances, typically including diagnoses, procedures, medications, and outcomes, but without trajectory modeling or demographic alignment. To highlight this contrast, both representations are shown in parallel.

This visual and textual comparison underscores the key design differences between our method and prior approaches, namely, our focus on severity-aware trajectories, explicit causal chains, and salient textual reasoning, versus the static, structured summaries used in the existing methods.

\section{Prompts}
\input{Prompt/prompt_severity.tex}
\input{Prompt/prompt_kst.tex}
\subsection{Prompt for Severity Score Generation}\label{appendix_Prompt_Severity}
As illustrated in Figure~\ref{fig:prompt_severity}, the prompt guides the LLM to assess the clinical severity of diagnosis concepts based on medical literature and guidelines, producing severity scores that are integrated into the severity-weighted medical knowledge graph. These scores enable the model to distinguish high-risk diagnoses from less critical ones during downstream reasoning.

\subsection{Prompts for Generating KSTs}\label{appendix_prompt_reconstruct}
As illustrated in Figure~\ref{fig:prompt_reconstruct}, the prompt enables the selection and refinement of key supporting trajectories by balancing relevance, severity, and diversity through natural language inference signals.

\subsection{Prompt for Inference}
As illustrated in Figure~\ref{fig:prompt_reason1} and Figure~\ref{fig:prompt_reason2}, the prompts integrate patient-specific trajectories, retrieved clinical notes, demographic information, and similar patient cases to support conservative and clinically grounded decision-making, while encouraging step-by-step reasoning and avoiding overconfident risk predictions.

\input{Prompt/prompt_mortality.tex}
\input{Prompt/prompt_readmission.tex}

\end{document}

%% file: macro.tex
\definecolor{darkgreen}{rgb}{0.0, 0.5, 0.0}

\newcommand{\vect}[1]{
  \mathbf{#1}
}

\newcommand{\miii}{MIMIC-III\xspace}
\newcommand{\miv}{MIMIC-IV\xspace}


%% file: Prompt/prompt_severity.tex
\begin{figure}[t]
\centering
\tcbset{colback=white, colframe=black!70, fonttitle=\bfseries}

\begin{tcolorbox}[colback=gray!5!white,colframe=gray!75!black, title=Severity Score Prompt]
\texttt{You are a board\textendash certified clinician. For each diagnosis provided, assign a severity score from 1 to 20, and return a single JSON object mapping diagnosis names to scores. Below are multiple diagnoses, each accompanied by ``PubMed Abstract'' and ``Wikipedia Summary'' as contextual information.\\
\\
For each diagnosis, output only the integer score (1--20) in a JSON mapping.\\
\\
\#\#\# INPUT  \\
**Diagnosis:** \\{diagnosis} \\
**PubMed Abstract:** \\{abstract} \\
**Wikipedia Summary:** \\{summary}
}
\end{tcolorbox}
\vspace{-10pt}
\caption{Prompt for Severity Score Generation.}
\label{fig:prompt_severity}
\end{figure}

%% file: Prompt/prompt_kst.tex
\begin{figure}[t]
\centering
\tcbset{colback=white, colframe=black!70, fonttitle=\bfseries}

\begin{tcolorbox}[colback=gray!5!white,colframe=gray!75!black, title=KST Retrieval Prompts]
\# Risk Path Query\par
Clinical deterioration. Retrieve trajectories showing worsening physiological conditions, progressive organ failure, or unresolved critical illness. Prioritize relations that indicate forward disease progression: [progresses\_to, results\_in, triggers, is\_complication\_of, contributes\_to, increases\_risk\_of, exacerbates, may\_cause, can\_lead\_to]. Avoid any relations suggesting recovery, resolution, or therapeutic benefit.\par\medskip
\# Protective Path Query\par
Clinical recovery. Retrieve trajectories showing stabilization after critical illness, physiological improvement, or reduced mortality risk. Focus on relations that reflect successful treatment or disease resolution: [resolves, improves, stabilizes, reduces\_risk\_of, ameliorates, leads\_to\_recovery\_from, is\_manageable\_with]. Avoid paths indicating deterioration or escalation.
\end{tcolorbox}
\vspace{-10pt}
\caption{Prompt for KST retrieval.}
\label{fig:prompt_reconstruct}
\end{figure}

%% file: Prompt/prompt_mortality.tex
\begin{figure*}[t]
\centering
\tcbset{colback=white, colframe=black!70, fonttitle=\bfseries}
\small
\begin{tcolorbox}[
  colback=gray!5!white,
  colframe=gray!75!black,
  title=Mortality Inference Prompt,
  width=\textwidth
]
\texttt{[Reasoning] Given the following task description, patient trajectory path, patient demographics, patient-specific clinical notes, and top-5 similar patients, please provide a step-by-step reasoning process that leads to the prediction outcome based on the patient’s context. After the reasoning process, provide the prediction label (0/1).\\
\\
\#\#\# Task \\
Mortality Prediction Task\\
Predict the mortality outcome for a patient’s subsequent hospital visit.\\
Labels: 0 = survival, 1 = mortality\\
\\
Regulator – Predict \texttt{"Survival (0)"} unless there is overwhelming evidence to predict \texttt{"Mortality (1)"}.\\
\\
Criteria for \texttt{Mortality (1)}:\\
- Critical high-risk condition with evidence of continuous deterioration\\
- Non-responsiveness to intensive care\\
\\
Do NOT predict mortality if:\\
- Mild-to-moderate deterioration\\
- Uncertainty or signs of recovery\\
\\
\#\#\# Clinical Assessment Summary \\
Protective Paths: \\{protective\_text} \\
Risk Paths: \\{risk\_text} \\
\\
\#\#\# Patient demographics \\
{demographics} \\
\\
\#\#\# Clinical Notes \\
{retrieved\_clinical\_notes} \\
\\
\#\#\# Similar Patients \\
{similar\_patients\_context} \\
\\
\#\#\# Step-by-Step Reasoning Guide \\
1. Review protective and risk paths\\
2. Determine severity and progression\\
3. Match against mortality criteria\\
4. Compare with similar patients\\
5. Weigh protective signals\\
6. Decide conservatively\\
\\
\#\#\# Output (Reasoning and Final prediction) \\
\# Reasoning \#\\
\\
\# Final prediction \#\\
(If any information of demographics is out of bounds, ignore it.)}
\end{tcolorbox}

\caption{Prompt for Mortality Risk Reasoning.}
\label{fig:prompt_reason1}
\end{figure*}

%% file: Prompt/prompt_readmission.tex
\begin{figure*}[t]
\centering
\tcbset{colback=white, colframe=black!70, fonttitle=\bfseries}
\small
\begin{tcolorbox}[
  colback=gray!5!white,
  colframe=gray!75!black,
  title=Readmission Prediction Prompt,
  width=\textwidth
]
\texttt{[Reasoning] Given the following task description, patient trajectory path, patient demographics, patient-specific clinical notes, and top-5 similar patients, please provide a step-by-step reasoning process that leads to the prediction outcome based on the patient’s context. After the reasoning process, provide the prediction label (0/1).\\
\\
\#\#\# Task \\
15-Day Readmission Prediction Task\\
Predict whether the patient will be readmitted within 15 days of discharge.\\
Labels: 0 = no readmission, 1 = readmission\\
\\
Regulator – Predict \texttt{"No Readmission (0)"} unless there is clear, overwhelming evidence of high risk.\\
\\
Criteria for \texttt{Readmission (1)}:\\
- Persistent or worsening symptoms despite treatment\\
- Poor discharge planning or lack of social support\\
- Critical unresolved issues (e.g., uncontrolled infection, CHF decompensation)\\
\\
Do NOT predict readmission if:\\
- Mild concerns, uncertainty, or adequate discharge planning\\
\\
\#\#\# Clinical Assessment Summary \\
Protective Paths: \\{protective\_text} \\
Risk Paths: \\{risk\_text} \\
\\
\#\#\# Patient demographics \\
{demographics} \\
\\
\#\#\# Clinical Notes \\
{retrieved\_clinical\_notes} \\
\\
\#\#\# Similar Patients \\
{similar\_patients\_context} \\
\\
\#\#\# Step-by-Step Reasoning Guide \\
1. Review protective vs. risk paths\\
2. Assess severity and trajectory\\
3. Match against readmission criteria\\
4. Contrast with similar patients\\
5. Evaluate discharge planning\\
6. Decide conservatively\\
\\
\#\#\# Output (Reasoning and Final Prediction) \\
\# Reasoning \#\\
\\
\# Final prediction \#
}
\end{tcolorbox}

\caption{Prompt for Readmission Risk Reasoning.}
\label{fig:prompt_reason2}
\end{figure*}